# Differential Privacy as a Perk: Federated Learning over Multiple-Access Fading Channels with a Multi-Antenna Base Station

Hao Liang, Haifeng Wen, Kaishun Wu, and Hong Xing

*Abstract*—Federated Learning (FL) is a distributed learning paradigm that preserves privacy by eliminating the need to exchange raw data during training. In its prototypical edge instantiation with underlying wireless transmissions enabled by analog over-the-air computing (AirComp), referred to as *over-the-air FL (AirFL)*, the inherent channel noise plays a unique role of *frenemy* in the sense that it degrades training due to noisy global aggregation while providing a natural source of randomness for privacy-preserving mechanisms, formally quantified by *differential privacy (DP)*. It remains, nevertheless, challenging to effectively harness such channel impairments, as prior arts, under assumptions of either simple channel models or restricted types of loss functions, mostly considering (local) DP enhancement with a single-round or non-convergent bound on privacy loss. In this paper, we study AirFL over multiple-access fading channels with a multi-antenna base station (BS) subject to user-level DP requirements. Despite a recent study, which claimed in similar settings that artificial noise (AN) must be injected to ensure DP in general, we demonstrate, on the contrary, that DP can be gained as a *perk* even *without* employing any AN. Specifically, we derive a novel bound on DP that converges under general bounded-domain assumptions on model parameters, along with a convergence bound with general smooth and non-convex loss functions. Next, we optimize over receive beamforming and power allocations to characterize the optimal convergence-privacy trade-offs, which also reveal explicit conditions in which DP is achievable without compromising training. Finally, our theoretical findings are validated by extensive numerical results.

## I. Introduction

FEDERATED learning (FL) has emerged as a promising distributed learning paradigm, enabling multiple devices to collaboratively train a shared model without exposing their local data. This protocol offers significant advantages in terms of data privacy and communication efficiency over traditional centralized approaches. In modern wireless networks, the rapid proliferation of intelligent edge devices—such as smartphones and IoT sensors—has resulted in massive volumes of data being generated at the edge, beyond the reach of centralized base stations (BSs). This trend has spurred growing interest in wireless FL. As envisioned in [1], wireless FL is expected to be a key enabler of flexible, scalable, and privacy-aware intelligence acquisition from numerous data-siloed entities.

H. Liang, H. Wen, K. Wu, and H. Xing are with the IoT Thrust, The Hong Kong University of Science and Technology (Guangzhou), Guangzhou, 511453, China; H. Xing is also affiliated with the Department of ECE, The Hong Kong University of Science and Technology, HK SAR (e-mails: hliang346@connect.hkust-gz.edu.cn, hwen904@connect.hkust-gz.edu.cn, wuks@hkust-gz.edu.cn, hongxing@ust.hk).

Over-the-air federated learning (AirFL) represents a paradigm shift in wireless FL by enabling multiple clients to simultaneously access the uplink channel through the superposition of radio-frequency (RF) waveforms, thereby achieving scalable training that remains communication-efficient regardless of the number of participating clients [2]–[4]. However, the practical deployment of AirFL faces significant challenges, including channel noise and fading inherent to analog transceivers, which deteriorate learning performance by introducing errors in model aggregation [5]. Numerous prior efforts have focused on analyzing and enhancing the convergence of AirFL, the key metric for training efficacy. For instance, Zhu *et al.* in [4] introduced a broadband AirFL scheme using truncated channel inversion, elucidating the trade-off between signal-to-noise ratio (SNR) and the expected update-truncation ratio. Building on this, a truncated channel inversion-based AirFL system with memory mechanisms was proposed by [6], achieving optimal convergence rates under fading channels. To further reduce the communication overhead, Zhu *et al.* in [7] proposed a one-bit over-the-air aggregation scheme integrated with AirFL while analyzing its convergence under noisy channels. Cao *et al.* in [8] derived and optimized transmission powers to minimize a convergence error upper bound of AirFL. Moreover, for an edge server equipped with a multi-antenna base station (BS), Yang *et al.* proposed multiple-input and multiple-output (MIMO)-based AirFL and maximized the number of joint devices by joint device selection and beamforming optimization subject to the aggregation error requirement [3]. When the number of antennas is large, [9] and [10] exploited channel hardening to achieve near-optimal aggregation with reduced pilot overhead and improved scalability.

On the other hand, naive privacy preserving by FL is not sufficient against emerging privacy attacks, such as membership inference attack [11], [12], and differential privacy (DP) is a formal performance metric with theoretical guarantee [13]–[16]. In federated settings, DP is typically implemented in one of two protocols: local DP (LDP) or user-level DP. In LDP, each client perturbs its local update before transmitting it to the server, thereby safeguarding individual data points and combating inference attacks on shared data values [17], [18]. In contrast, user-level DP, which is more commonly adopted in FL, aims to protect a user's entire contribution throughout the FL training process [19], [20]. This is typically achieved by allowing a trusted server to aggregate the parameter updates from multiple users and then applying a calibrated noise mechanism to the aggregated result [21].

The most important class of mechanisms to achieve DP is the Gaussian mechanism, by which the calibrated Gaussian noise is added to the output of a function to mask the contribution of any input data. In this regard, the above-mentioned channel noise in AirFL due to wireless analog transmission is nonetheless beneficial to DP, thus becoming a frenemy in terms of privacy-preserving training. There was previous work that investigated DP employing artificial noise alongside the intrinsic channel noise in wireless FL settings. For instance, Liu and Simeone in [22] considered conditions in which DP is guaranteed without compromising training, i.e., *for free*. Subsequently, Hu *et al.* in [23] proposed an exploration into device sampling policies with replacement as a potential mechanism for augmenting the DP levels of devices by assuming that inactive devices can choose to transmit artificial noise. Also, Park and Choi in [24] considered the inherent randomness of the local gradient, which can be used to enhance the privacy analysis. However, [23], [24] only considered single-antenna BS over (quasi-) static AWGN channels, which cannot employ spatial diversity to improve training-privacy trade-offs. Although [25], [26] studied multi-antenna BS over possibly more complex channels, the analysis of convergence bounds was built upon assumptions on the loss function satisfying Polyak-Lojasiewicz conditions or being strongly convex. In addition, [27], [28] optimized the power allocation policy leveraging the channel noise, but focused on privacy loss in a single round, failing to capture the effect of mechanism composition spanning over the entire training.

To gain a fundamental understanding of the role played by channel noise in DP preserving, in this paper, we consider a practical AirFL setting where multiple wireless devices (WDs) collaborate to train or fine-tune a shared model by uplink transmissions over multiple-access fading channels. We are typically interested in answering what is the optimal receive beamforming design for the multi-antenna BS to achieve convergence-privacy trade-offs with general smooth and non-convex loss functions, and whether or not there is any DP gained as a perk if we only aim for convergence improvement. Recently, Liu *et al.* proposed a transceiver design in a similar setup under the assumption of full client participation [26]. However, their approach suffers from several technical limitations. First, their framework provides a non-convergent, sample-level DP guarantee, which is less practical for the FL setting and offers limited privacy guarantees when the number of communication rounds becomes large. Besides, their convergence analysis is restricted to the case with strongly-convex loss functions, thus challenging its theoretical validity in modern deep learning applications. Moreover, the effect of gradient clipping is not characterized, which is an essential component for guaranteeing DP. Above all, they asserted that "privacy-for-free" is not generally achievable for AirFL in such settings, and hence (additional) artificial noise mechanisms at device side must be required, a notion that our work challenges. The main contributions of the paper are summarized as follows:

- A convergent upper bound on DP is derived under general smooth and non-convex loss functions with a bounded parameter domain assumption. It does not grow infinitely with the number of communication rounds after a burn-in period, thus offering a much tighter and more practical privacy characterization.
- We formulate and solve an optimization problem to jointly design the receive beamforming and power allocation. Our design achieves an optimal trade-off between model convergence and the user-level DP guarantee.
- Our analysis explicitly reveals that the zero-artificial-noise property always holds for general multi-antenna cases. We also reveal explicit conditions in which DP is gained as a perk in AirFL with no influence in training.
- The numerical experiments demonstrate that in practical settings, benchmarking AirFL training without privacy concerns, DP can be gained as a perk starting from a reasonably high privacy requirement and in low signal-to-noise ratio (SNR) regimes, beyond a comparable level of test accuracy.

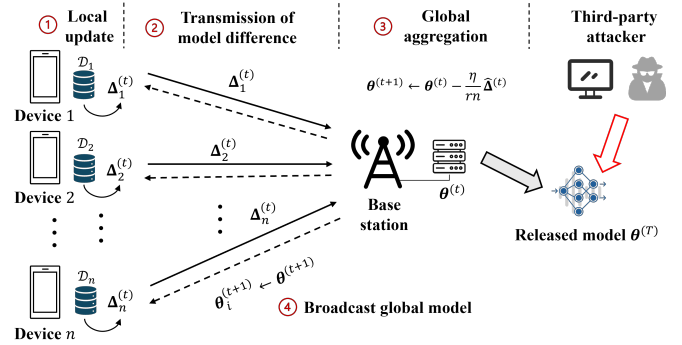

Fig. 1. An overview of the federated learning framework under the considered threat model.

The remainder of this paper is organized as follows. The system model and problem formulation are presented in Sec. II. Privacy analysis is provided in Sec. III. The convergence analysis and optimal transceiver design policies are presented in Sec. IV. Experiments are performed to validate the analyses in Sec. V, followed by conclusions and discussions in Sec. VI.

*Notation*—We use the upper case boldface letters for matrices and lower case boldface letters for vectors. We also use $\|\cdot\|$ to denote the Euclidean norm of a vector. Notations $(\cdot)^T$ and $(\cdot)^H$ denote the transpose and the conjugate transpose of a matrix, respectively. $\mathbb{E}[\cdot]$ stands for the statistical expectation of a random variable. $\boldsymbol{I}_d$ denotes a $d$-dimensional identity matrix, and $\triangleq$ indicates a mathematical definition. We denote by $\Pr[\cdot]$ the probability of a random event. The law of a random variable $\boldsymbol{\mu}$ is denoted as $\mathbb{P}_{\boldsymbol{\mu}}$. $|\mathcal{X}|$ denotes the cardinality of the set $\mathcal{X}$.

## II. SYSTEM MODEL AND PROBLEM FORMULATION

In this section, we delineate the system model and the foundational concepts essential for our analysis. Throughout this work, we study a wireless federated learning system comprising an $m$-antenna BS and $n$ single-antenna WDs connected through it via a multiple access (MAC) fading channel, as shown in Fig. 1.

## A. Learning Protocol (Vanilla FL)

The WDs and the BS cooperatively solve the empirical loss minimization problem, defined as

$$\min_{\boldsymbol{\theta} \in \mathbb{R}^d} f(\boldsymbol{\theta}) \triangleq \frac{1}{n} \sum_{i=1}^{n} f_i(\boldsymbol{\theta}), \quad (1)$$

where $f_i(\boldsymbol{\theta}) \triangleq (1/|\mathcal{D}_i|) \sum_{\xi \in \mathcal{D}_i} \ell(\boldsymbol{\theta}; \xi)$ is the local empirical loss function at WD $i \in [n]$; $\mathcal{D}_i$ denoting the local dataset at WD $i \in [n]$; and $\ell(\boldsymbol{\theta}; \xi)$ denotes the loss function of a model parameterized by $\boldsymbol{\theta} \in \mathbb{R}^d$ evaluated on the data sample $\xi$.

Federated averaging (FedAvg) provides an efficient way to solve (1) in a distributed manner [19], termed as the *vanilla FL* protocol throughout this paper. Denote the index of a communication round for global aggregation by $t \in \{0, \ldots, T-1\}$ and the active device set at round $t$ by $\mathcal{I}^{(t)}$. At the $t$-th global communication round, the BS uniformly selects $|\mathcal{I}^{(t)}| = rn$ active devices at random, $r \in (0, 1]$, and broadcasts the current global model parameter $\boldsymbol{\theta}^{(t)}$ to all WDs as initialization of the local models.

Next, the active WD $i \in \mathcal{I}^{(t)}$ performs $Q$ local stochastic gradient descent (SGD) steps on its own dataset $\mathcal{D}_i$ using the initialized model $\boldsymbol{\theta}_i^{(t,0)} = \boldsymbol{\theta}^{(t)}$, resulting in

$$\textbf{(Local Update)} \quad \boldsymbol{\theta}_i^{(t,q+1)} = \boldsymbol{\theta}_i^{(t,q)} - \eta \hat{\nabla} f_i(\boldsymbol{\theta}_i^{(t,q)}), \quad (2)$$

where $q \in \{0, \ldots, Q-1\}$ denotes the local iteration index; $\eta$ is the local learning rate; and $\hat{\nabla} f_i(\boldsymbol{\theta}_i^{(t,q)})$ is the stochastic gradient obtained from a mini-batch $\mathcal{B}_i^{(t,q)} \subseteq \mathcal{D}_i$ of the local dataset, i.e.,

$$\hat{\nabla} f_i(\boldsymbol{\theta}_i^{(t,q)}) = \frac{1}{|\mathcal{B}_i^{(t,q)}|} \sum_{\xi \in \mathcal{B}_i^{(t,q)}} \nabla \ell(\boldsymbol{\theta}_i^{(t,q)}; \xi). \quad (3)$$

Then, each WD $i \in \mathcal{I}^{(t)}$ transmits the (scaled) model difference

$$\boldsymbol{\Delta}_i^{(t)} = \frac{1}{\eta} \left( \boldsymbol{\theta}_i^{(t,0)} - \boldsymbol{\theta}_i^{(t,Q)} \right) \quad (4)$$

to the BS. We will elaborate on the transmission schemes in the next subsection.

Finally, the server updates the global model parameters, $\boldsymbol{\theta}^{(t+1)}$, by performing averaging as follows

$$\textbf{(Global Aggregation)} \quad \boldsymbol{\theta}^{(t+1)} = \boldsymbol{\theta}^{(t)} - \frac{\eta}{rn} \sum_{i \in \mathcal{I}^{(t)}} \boldsymbol{\Delta}_i^{(t)}. \quad (5)$$

The above steps iterate until a suitable convergence criterion is satisfied.

## B. Communication Model

In this paper, we study a wireless system in which all WDs communicate with the BS over MAC fading channels via over-the-air computing (AirComp) [2].

At the $t$-th global communication round, each active WD $i$ in $\mathcal{I}^{(t)}$ transmits the clipped version of the model difference to the server, which can be expressed as

$$\boldsymbol{x}_i^{(t)} = s_i^{(t)} \text{clip}_c \left( \boldsymbol{\Delta}_i^{(t)} \right), \quad (6)$$

where $\text{clip}_c(\boldsymbol{x}) = \boldsymbol{x} \cdot \min(1, c/\|\boldsymbol{x}\|)$ is the clipping operator and $c \in \mathbb{R}^+$ is the clipping threshold given as a hyper parameter; and $s_i^{(t)} \in \mathbb{C}$ is a power scaling factor designed to satisfy the transmit power constraint averaging over $d$ symbols, as

$$\frac{1}{d} \mathbb{E}\left[ \|\boldsymbol{x}_i^{(t)}\|^2 \right] \leq P. \quad (7)$$

Note that clipping not only facilitates instantaneous power control at low-power WDs but also ensures DP, which will be revealed in the next section.

We assume a block flat-fading channel $\boldsymbol{h}_i^{(t)}$, where the channel coefficients remain constant within a communication block, but may vary from block to block. Also, we consider symbol-level synchronization among the devices that transmit each entry of $\boldsymbol{x}_i^{(t)}$, $x_{i,j}$ for $j \in [d]$, simultaneously. As a result, at the $t$-th round, the $j$-th entry of the received signal vector at the BS for the $j$-th entry, $\boldsymbol{y}_j^{(t)}$, is given by

$$\begin{aligned} \boldsymbol{y}_j^{(t)} &= \sum_{i \in \mathcal{I}^{(t)}} \boldsymbol{h}_i^{(t)} x_{i,j}^{(t)} + \boldsymbol{n}_j^{(t)} \\ &= \sum_{i \in \mathcal{I}^{(t)}} \boldsymbol{h}_i^{(t)} s_i^{(t)} \bar{\Delta}_{i,j}^{(t)} + \boldsymbol{n}_j^{(t)}, \end{aligned} \quad (8)$$

where $\bar{\boldsymbol{\Delta}}_i^{(t)} = \text{clip}_c(\boldsymbol{\Delta}_i^{(t)})$ denotes the scaling model difference clipped by the $i$-th device; and $\boldsymbol{n}_j^{(t)} \in \mathbb{C}^m$ is the circular symmetric complex Gaussian (CSCG) noise received at the BS, denoted by $\boldsymbol{n}_j^{(t)} \sim \mathcal{CN}(\boldsymbol{0}, \sigma^2 \boldsymbol{I}_m)$.

Subsequently, the BS computes $\hat{\boldsymbol{\Delta}}^{(t)}$ as an estimate of the sum of model differences $\sum_{i \in \mathcal{I}^{(t)}} \boldsymbol{\Delta}_i^{(t)}$ by a receive combiner $\boldsymbol{w}^{(t)}$, which yields

$$\begin{aligned} \hat{\Delta}_j^{(t)} &= (\boldsymbol{w}^{(t)})^H \boldsymbol{y}_j^{(t)} \\ &= \sum_{i \in \mathcal{I}^{(t)}} (\boldsymbol{w}^{(t)})^H \boldsymbol{h}_i^{(t)} s_i^{(t)} \bar{\Delta}_{i,j}^{(t)} + (\boldsymbol{w}_j^{(t)})^H \boldsymbol{n}_j^{(t)}, \end{aligned} \quad (9)$$

where $\hat{\Delta}_j^{(t)}$ denotes the $j$-th entry of $\hat{\boldsymbol{\Delta}}^{(t)}$, $j \in [d]$. Note that we employ the same combiner $\boldsymbol{w}^{(t)}$ to estimate all entries of $\hat{\boldsymbol{\Delta}}^{(t)}$ during the same communication round, the optimality of which will be shown in Sec. IV-C. Upon stacking $\hat{\boldsymbol{\Delta}}^{(t)}$, the BS updates the global model as

$$\boldsymbol{\theta}^{(t+1)} = \boldsymbol{\theta}^{(t)} - \frac{\eta}{rn} \hat{\boldsymbol{\Delta}}^{(t)}. \quad (10)$$

As in [22], [26], [29], we assume that the downlink communication is ideal and noiseless, allowing each device to receive the global model $\boldsymbol{\theta}^{(t)}$ from the BS without distortion. This assumption is justified because the BS typically has far fewer resource constraints compared to the WDs for uplink transmissions. We refer to the above algorithm as *AirFL-DP*, which is summarized in Algorithm 1.

## C. Privacy Model

As shown in Fig. 1, we consider that there is a *curious* third-party attacker who particularly attempts to infer information related to certain users (WDs), thus compromising *user-level privacy*, while the WDs and the BS are assumed to be trustful. As in [30]–[32], we assume a black-box setting where the attacker can only access the system's final-released model, $\boldsymbol{\theta}^{(T)}$, after the entire FL process is complete without access to

## Algorithm 1: AirFL-DP

1 **Input:** learning rate $\eta$, power constraints $P$, number of communication rounds $T$, number of local rounds $Q$, active device ratio $r$, and clipping threshold $c$
2 Initialize $\boldsymbol{\theta}_i^{(0)} = \boldsymbol{\theta}^{(0)}$
3 **while** $t < T$ **do**
4     **On devices** $i \in \mathcal{I}^{(t)}$ **(in parallel):**
5         $\boldsymbol{\theta}_i^{(t,0)} \leftarrow \boldsymbol{\theta}^{(t)}$;
6         **for** $q = 0$ **to** $Q-1$ **do**
7             $\boldsymbol{\theta}_i^{(t,q+1)} = \boldsymbol{\theta}_i^{(t,q)} - \eta \hat{\nabla} f_i(\boldsymbol{\theta}_i^{(t,q)})$;
8         **end**
9         $\boldsymbol{\Delta}_i^{(t)} = (\boldsymbol{\theta}_i^{(t,0)} - \boldsymbol{\theta}_i^{(t,Q)})/\eta$;
10         Transmit $\boldsymbol{x}_i^{(t)} = s_i^{(t)} \text{clip}_c(\boldsymbol{\Delta}_i^{(t)})$;
11     **end**
12     **On server:**
13         Receive $\boldsymbol{y}_j^{(t)} = \sum_{i \in \mathcal{I}^{(t)}} \boldsymbol{h}_i^{(t)} x_{i,j}^{(t)} + \boldsymbol{n}_j^{(t)}$ for $j \in [d]$;
14         Combining: $\hat{\Delta}_j^{(t)} = (\boldsymbol{w}^{(t)})^H \boldsymbol{y}_j^{(t)}$ for $j \in [d]$;
15         Global update: $\boldsymbol{\theta}^{(t+1)} = \boldsymbol{\theta}^{(t)} - \frac{\eta}{rn} \hat{\boldsymbol{\Delta}}^{(t)}$;
16         Broadcast $\boldsymbol{\theta}^{(t+1)}$ to all $n$ devices;
17     **end**
18     $t \leftarrow t + 1$;
19 **end**
20 **Output:** $\boldsymbol{\theta}^{(T)}$

any intermediate training artifacts (e.g., gradients or parameter updates).

We begin formally quantifying the above privacy by recalling the definition of DP, which provides a standard framework to ensure that a learning model's output remains almost unchanged when applied to two *user-adjacent datasets*, thus leading to user-level DP. Rather than protecting privacy for single examples as previous works [22], [26], the user-adjacent datasets lead to formal guarantees of user-level privacy.

*Definition 2.1 (User-Adjacent Datasets [16]):* The datasets $\mathcal{D}$ and $\mathcal{D}'$ are said to be *user-adjacent* if one can be obtained from the other by adding or removing all samples associated with that user. Formally, $\mathcal{D}$ and $\mathcal{D}'$ satisfy that $\mathcal{D}' = \mathcal{D} \cup \mathcal{D}_{i^*}$ (set addition) or $\mathcal{D}' = \mathcal{D} \setminus \mathcal{D}_{i^*}$ (set removal), where $\mathcal{D}_{i^*}$ denotes the set of all samples associated with a particular user $i^*$.

*Definition 2.2 (Differential Privacy [13]):* For $\epsilon \geq 0$, $\delta \in [0,1]$, a randomized mechanism $\mathcal{M}: 2^\mathcal{X} \mapsto \mathcal{Y}$ is $(\epsilon, \delta)$-DP if, for every pair of user-adjacent datasets, $\mathcal{D}, \mathcal{D}' \subseteq \mathcal{X}$, and for any subset of outputs $\mathcal{S} \subseteq \mathcal{Y}$, we have

$$\Pr[\mathcal{M}(\mathcal{D}) \in \mathcal{S}] \leq \exp(\epsilon) \Pr[\mathcal{M}(\mathcal{D}') \in \mathcal{S}] + \delta. \quad (11)$$

## III. PRIVACY ANALYSIS

In this section, we present a user-level DP analysis of the AirFL-DP algorithm described in Sec. II, demonstrating that AirFL-DP achieves user-level $(\epsilon, \delta)$-DP guarantee for any $\epsilon > 0$ without additional artificial noise. Moreover, our analysis shows that the accumulated privacy loss is bounded by a constant independent of the number of rounds $T$, which, to the best of our knowledge, has been presented for the first time in the AirFL literature.

### A. Preliminaries

Our analysis is based on the privacy amplification by iteration framework [30] leveraging Rényi differential privacy (RDP) [33], which facilitates tracking privacy loss and obtaining a tighter privacy bound than standard DP analytic tools such as strong composition [34] and moment accounting [15].

*Definition 3.1 (User-Level Rényi Differential Privacy [33]):* For $\alpha > 1$, $\epsilon' \geq 0$, a randomized mechanism $\mathcal{M}: 2^\mathcal{X} \mapsto \mathcal{Y}$ satisfies $(\alpha, \epsilon')$-RDP if, for any pair of user-adjacent datasets, $\mathcal{D}, \mathcal{D}' \subseteq \mathcal{X}$, it holds that

$$D_\alpha(\mathbb{P}_{\mathcal{M}(\mathcal{D})} || \mathbb{P}_{\mathcal{M}(\mathcal{D}')}) \leq \epsilon', \quad (12)$$

where $D_\alpha(\mathbb{P}_\mu || \mathbb{P}_{\mu'})$ is Rényi divergence defined as [35]

$$D_\alpha(\mathbb{P}_\mu || \mathbb{P}_{\mu'}) = \frac{1}{\alpha - 1} \log \mathbb{E}_{s \sim \mathbb{P}_{\mu'}} \left\{ \left( \frac{\Pr[\mu = s]}{\Pr[\mu' = s]} \right)^\alpha \right\}. \quad (13)$$

Note that RDP can be easily transformed into an equivalent characterization of DP via the following lemma.

*Lemma 3.1 (From $(\alpha, \epsilon')$-RDP to $(\epsilon, \delta)$-DP [33]):* If $\mathcal{M}$ is an $(\alpha, \epsilon')$-RDP mechanism, it also satisfies $(\epsilon' + \frac{\log 1/\delta}{\alpha - 1}, \delta)$-DP for any $0 < \delta < 1$.

Next, we introduce a definition that plays a key role in the framework of the privacy amplification by iteration [30].

*Definition 3.2 (Shifted Rényi Divergence [30]):* Let $\boldsymbol{\mu}$ and $\boldsymbol{\nu}$ be two random variables with distributions $\mathbb{P}_{\boldsymbol{\mu}}$ and $\mathbb{P}_{\boldsymbol{\nu}}$. For any shift parameter $z \geq 0$ and Rényi order $\alpha > 1$, the $z$-shifted Rényi divergence is defined as

$$\mathcal{D}_\alpha^{(z)}(\mathbb{P}_{\boldsymbol{\mu}} || \mathbb{P}_{\boldsymbol{\nu}}) = \inf_{\mathbb{P}_{\boldsymbol{\mu}'}: W_\infty(\mathbb{P}_{\boldsymbol{\mu}}, \mathbb{P}_{\boldsymbol{\mu}'}) \leq z} \mathcal{D}_\alpha(\mathbb{P}_{\boldsymbol{\mu}'} || \mathbb{P}_{\boldsymbol{\nu}}), \quad (14)$$

where $W_\infty(\cdot, \cdot)$ denotes the $\infty$-Wasserstein distance.

### B. User-level Privacy Guarantee

We adopt the following assumptions throughout our privacy analysis.

*Assumption 1 (L-Smoothness):* The loss function $\ell(\cdot; \xi)$ is smooth with constant $L > 0$, i.e., for any $\boldsymbol{\theta}, \boldsymbol{\theta}' \in \mathbb{R}^d$,

$$\left\| \nabla \ell(\boldsymbol{\theta}; \xi) - \nabla \ell(\boldsymbol{\theta}'; \xi) \right\| \leq L \left\| \boldsymbol{\theta} - \boldsymbol{\theta}' \right\|. \quad (15)$$

*Assumption 2 (Bounded Parameter Domain):* For any iteration $t \in \{0, \ldots, T-1\}$, the domain of the model parameters has diameter $0 < D < \infty$.

Assumption 2 is a practical and mild condition in optimization. Despite of being not strictly necessary, it enables tighter and convergent privacy bounds. This assumption is well-justified, as many optimization problems operate within constrained parameter spaces due to problem structure or practical constraints. For unconstrained problems, a sequence of constrained subproblems can be solved with negligible computational and privacy overhead [36].

The privacy analysis begins with the global update rule (10), which is rewritten as:

$$\boldsymbol{\theta}^{(t+1)} = \boldsymbol{\theta}^{(t)} - \frac{\eta}{rn} \sum_{i \in \mathcal{I}^{(t)}} (\boldsymbol{w}^{(t)})^H \boldsymbol{h}_i^{(t)} s_i^{(t)} \bar{\boldsymbol{\Delta}}_i^{(t)} + \boldsymbol{n}_\text{eq}^{(t)}, \quad (16)$$

where $\boldsymbol{n}_{\text{eq}}^{(t)} \sim \mathcal{CN}(\boldsymbol{0}, (\eta^2\|\boldsymbol{w}^{(t)}\|^2\sigma^2/(r^2n^2))\boldsymbol{I}_d)$ denotes the effective noise at iteration $t$. To facilitate the analysis, we split the effective noise, $\boldsymbol{n}_{\text{eq}}^{(t)}$, into two terms as $\boldsymbol{n}_{\text{eq}}^{(t)} = \boldsymbol{\varrho}^{(t)} + \boldsymbol{\varsigma}^{(t)}$, where

$$\boldsymbol{\varrho}^{(t)} \sim \mathcal{CN}\left(\boldsymbol{0}, \beta^{(t)}\eta^2\|\boldsymbol{w}^{(t)}\|^2\sigma^2/(r^2n^2)\boldsymbol{I}_d\right), \quad (17)$$
$$\boldsymbol{\varsigma}^{(t)} \sim \mathcal{CN}\left(\boldsymbol{0}, (1-\beta^{(t)})\eta^2\|\boldsymbol{w}^{(t)}\|^2\sigma^2/(r^2n^2)\boldsymbol{I}_d\right),$$

where $\beta^{(t)} \in [0,1]$ is a predefined constant.

The right-hand side (RHS) of (17) with $\boldsymbol{n}_{\text{eq}}^{(t)}$ involving only $\boldsymbol{\varrho}^{(t)}$ defines a *noisy update function* $\psi^{(t)}(\cdot): \mathbb{R}^d \mapsto \mathbb{R}^d$ as

$$\psi^{(t)}(\boldsymbol{\theta}^{(t)}) = \boldsymbol{\theta}^{(t)} - \frac{\eta}{rn}\sum_{i\in\mathcal{I}^{(t)}}(\boldsymbol{w}^{(t)})^H\boldsymbol{h}_i^{(t)}s_i^{(t)}\bar{\boldsymbol{\Delta}}_i^{(t)} + \boldsymbol{\varrho}^{(t)}. \quad (18)$$

The remaining term in the RHS of (16), with $\boldsymbol{\varsigma}^{(t)}$, is aimed for further bounding the shifted Rényi divergence leveraging the following lemma [30, Lemma 20].

*Lemma 3.2 (Shift-Reduction [30]):* Let $\boldsymbol{\mu}$, $\boldsymbol{\nu}$ be two $d$-dimensional random variables. Then, for any $a \geq 0$ and $z \geq 0$, we have

$$\mathcal{D}_\alpha^{(z)}(\mathbb{P}_{\boldsymbol{\mu}} * \mathbb{P}_{\boldsymbol{\zeta}} \| \mathbb{P}_{\boldsymbol{\nu}} * \mathbb{P}_{\boldsymbol{\zeta}}) \leq \mathcal{D}_\alpha^{(z+a)}(\mathbb{P}_{\boldsymbol{\mu}} \| \mathbb{P}_{\boldsymbol{\nu}}) + \frac{\alpha a^2}{2\sigma^2}, \quad (19)$$

where $\boldsymbol{\zeta} \sim \mathcal{N}(\boldsymbol{0}, \sigma^2\boldsymbol{I}_d)$, and $\mathbb{P}_{\boldsymbol{\mu}} * \mathbb{P}_{\boldsymbol{\zeta}}$ denotes the distribution of the sum $\boldsymbol{\mu} + \boldsymbol{\zeta}$ with $\boldsymbol{\mu}$ and $\boldsymbol{\zeta}$ drawn independently.

The privacy cost of $\psi^{(t)}(\cdot)$ is quantified via the following lemma.

*Lemma 3.3 (Noisy Smooth-Reduction):* Let $\psi^{(t)}$ and $\psi^{(t)\prime}$ be two noisy update functions of AirFL-DP based on user-adjacent datasets $\mathcal{D}$ and $\mathcal{D}'$ at iteration $t$, respectively. Under Assumption 1, for any random variables $\boldsymbol{\mu}$ and $\boldsymbol{\nu}$, we have

$$\mathcal{D}_\alpha^{((1+\kappa^{(t)})^Qz)}(\mathbb{P}_{\psi^{(t)}(\boldsymbol{\mu})} \| \mathbb{P}_{\psi^{(t)\prime}(\boldsymbol{\nu})})$$
$$\leq \mathcal{D}_\alpha^{(z)}(\mathbb{P}_{\boldsymbol{\mu}} \| \mathbb{P}_{\boldsymbol{\nu}}) + \frac{2\alpha rc^2 \max_{i\in\mathcal{I}^{(t)}} |(\boldsymbol{w}^{(t)})^H\boldsymbol{h}_i^{(t)}s_i^{(t)}|^2}{\beta^{(t)}\|\boldsymbol{w}^{(t)}\|^2\sigma^2}, \quad (20)$$

where $\kappa^{(t)} = \eta L/(rn)\sum_{i\in\mathcal{I}^{(t)}}(\boldsymbol{w}^{(t)})^H\boldsymbol{h}_i^{(t)}s_i^{(t)}$ and the randomness comes from SGD and channel noise.

*Proof:* The detailed proof is provided in Appendix B. ∎

We are now ready to present our main results.

*Proposition 3.1 (($\alpha, \epsilon'$)-RDP Guarantee for AirFL-DP):* Under Assumption 1 and 2, for any $\alpha > 1$, AirFL-DP satisfies user-level $(\alpha, \epsilon')$-RDP, where

$$\epsilon' = \frac{2\alpha rc^2}{\sigma^2}\min\left\{\sum_{t=0}^{T-1}\phi_t, \Phi\right\} \quad (21)$$

with

$$\phi_t = \frac{\max_{i\in\mathcal{I}^{(t)}}|(\boldsymbol{w}^{(t)})^H\boldsymbol{h}_i^{(t)}s_i^{(t)}|^2}{\|\boldsymbol{w}^{(t)}\|^2},$$
$$\Phi = \left(\sqrt{\phi_{T-1}} + \frac{(1+\kappa_{\max})^Q\sqrt{r}Dn}{2\eta c\|\boldsymbol{w}^{(T-1)}\|}\right)^2, \quad (22)$$

where $\kappa_{\max} = \max_t \eta L/(rn)\sum_{i\in\mathcal{I}^{(t)}}(\boldsymbol{w}^{(t)})^H\boldsymbol{h}_i^{(t)}s_i^{(t)}$.

*Proof:* We summarize the sketch of proof as follows. This proof proceeds by a recursion on the iteration index $t$, mainly involving the following key steps:

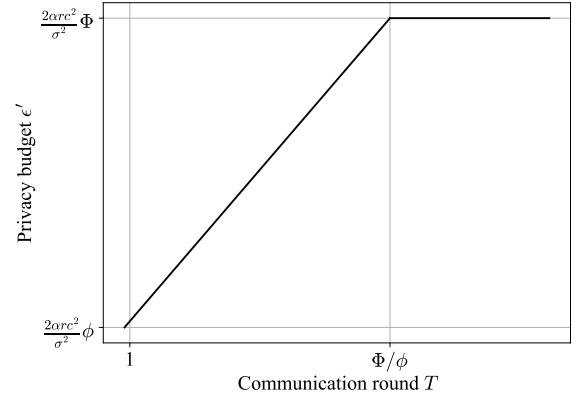

Fig. 2. An illustration of the RDP privacy loss $\epsilon'$ as a function of the total number of communication rounds $T$. For ease of exposition, this plot sets a constant $\phi_t = \phi$ for all $t$. The full analysis and parameter settings used in our experiments are specified in Section V-A.

1) **Decomposition**: At each iteration $t$, we conceptually split the AirFL-DP update procedure into two components: (i) a noisy update function and (ii) a shift-reduction mechanism that incorporates the remaining Gaussian noise.
2) **Base case** ($t = \tau$, for any $\tau \in \{0, 1, \ldots, T-1\}$): We leverage Assumption 2.
3) **Recursive step**: As for the recursive step, we first bound the privacy cost of the noisy update component using Lemma 3.3. Next, we analyze the shift-reduction mechanism using Lemma 3.2, which involves introducing auxiliary shift variables to control the shift parameter.
4) **Trace RDP loss**: The overall upper bound on RDP is derived by repeatedly applying the recursive step from $T$ to $\tau$ and appropriately setting the values of the auxiliary shift variables.

The detailed proof is provided in Appendix C. ∎

As shown in Fig. 2, a key feature of our analysis result is that the privacy loss can converge to a constant value irrespective of the number of communication rounds. In other words, the AirFL-DP algorithm will stop accumulating privacy loss after a sufficient number of iterations. This marks a significant improvement compared to previous analyses, where the privacy bound typically grows unboundedly with $T$ [22, Lemma 3], [26, Corollary 1]. Furthermore, our analysis gracefully recovers the linear bound when the domain diameter $D$ approaches infinity.

Another important insight gained from our analysis is the critical role played by the wireless channel impairments. Our proposition demonstrates that the DP cost is paid by inherent channel noise for free. It suggests that an arbitrary level $\epsilon$ is achievable by controlling the receive beamforming vector $\boldsymbol{w}^{(t)}$, power scaling factor $s_i^{(t)}$, and the clipping threshold $c$ in conditions of fading channel $\boldsymbol{h}_i^{(t)}$ and channel noise of power $\sigma^2$, $i \in \mathcal{I}^{(t)}$.

According to Lemma 3.1, a standard $(\epsilon, \delta)$-DP bound for AirFL-DP immediately follows, as a corollary of Proposition 3.1.

*Corollary 3.1 (($\epsilon, \delta$)-DP Guarantee for AirFL-DP):* Under Assumption 1 and 2, AirFL-DP with achievable $(\alpha, \epsilon')$-RDP

also satisfies user-level $(\epsilon, \delta)$-DP, for any $\delta \in (0, 1)$, with $\epsilon$ given by

$$\epsilon = \sqrt{\frac{(2c_\delta + 8)\log(1/\delta)rc^2}{\sigma^2} \min\left\{\sum_{t=0}^{T-1} \phi_t, \Phi\right\}}, \quad (23)$$

where $c_\delta \geq 4\epsilon'/\log(1/\delta)$, and $\phi_t$ and $\Phi$ are given in (22).
*Proof:* The detailed proof is provided in Appendix D. ∎

The privacy bound given by (23) in Corollary 3.1, jointly decided by the receive beamforming vector $\boldsymbol{w}^{(t)}$ and power scaling factor $s_i^{(t)}$ will be used to constrain the achievable DP level for AirFL-DP in next section.

## IV. OPTIMAL RECEIVE BEAMFORMING FOR FREE-DP AIRFL

In this section, we derive optimal receive beamforming designs to maximize learning performance, as characterized by the convergence bound for AirFL-DP training, subject to DP and transmit power constraints. To this end, we first perform the convergence analysis for AirFL-DP in Sec. IV-A, and then formulate the learning performance maximization problem in Sec. IV-B, followed by the optimal solution provided in Sec. IV-C.

### A. Convergence Analysis

We adopt the following standard assumptions from the FL literature.

*Assumption 3 (G-Lipschitz):* For any $\xi \in \mathcal{D}$ and all $\boldsymbol{\theta}, \boldsymbol{\theta}' \in \mathbb{R}^d$, there exists $G > 0$ such that

$$\left\|\ell(\boldsymbol{\theta}; \xi) - \ell(\boldsymbol{\theta}'; \xi)\right\| \leq G \left\|\boldsymbol{\theta} - \boldsymbol{\theta}'\right\|. \quad (24)$$

*Assumption 4 (Bounded SGD Variance and Norm):* For any $i \in [n]$ and $\boldsymbol{\theta} \in \mathbb{R}^d$, the stochastic batch gradient $\hat{\nabla} f_i(\cdot)$ has bounded variance $\sigma_l^2$, i.e.,

$$\mathbb{E}\left[\|\hat{\nabla} f_i(\boldsymbol{\theta}) - \nabla f_i(\boldsymbol{\theta})\|^2\right] \leq \sigma_l^2,$$

where the expectation is taken over the randomness of mini-batch selection.

*Assumption 5 (Gradient Dissimilarity):* For any $i \in [n]$ and $\boldsymbol{\theta} \in \mathbb{R}^d$, the gradient of local empirical loss function, $\nabla f_i(\cdot)$, satisfies the following inequality

$$\|\nabla f_i(\boldsymbol{\theta}) - \nabla f(\boldsymbol{\theta})\|^2 \leq \sigma_g^2. \quad (25)$$

Our convergence analysis builds on a recent result in [5], which addresses the convergence of our algorithm in a simplified setting. Specifically, they considered additive white Gaussian noise (AWGN) channel (i.e., $\boldsymbol{h}_i^{(t)} = \boldsymbol{1}$, for all $i \in [n]$ and $t$), full client participation ($r = 1$), and a single-antenna base station ($m = 1$). Their analysis relies on the assumption that the variance of estimation error $\boldsymbol{n}_{\text{est}}^{(t)} = \hat{\boldsymbol{\Delta}}^{(t)} - \sum_{i \in \mathcal{I}} \bar{\boldsymbol{\Delta}}_i^{(t)}$ can be expressed as

$$\mathbb{E}\left[\|\boldsymbol{n}_{\text{est}}^{(t)}\|^2\right] = dv^{(t)}, \quad (26)$$

where $v^{(t)} > 0$ is uncorrelated with the transmit signal, and numerically evaluable, while we provide upper bound on the mean-square error in (26), which explicitly depends on receive beamforming vector $\boldsymbol{w}^{(t)}$ and power scaling factor $s_i^{(t)}$, $i \in \mathcal{I}^{(t)}$, at each communication round, thus facilitating optimization problem formulation elaborated shortly.

Specifically, we adapt their convergence result in the following lemma.

*Lemma 4.1 (Theorem 4.1 in [5]):* Under the simplified setting above and Assumptions 1, 3–5, AirFL-DP with $\eta = \mathcal{O}(1/(QL))$ after $T$ rounds satisfies

$$\frac{1}{T}\sum_{t=0}^{T-1} \mathbb{E}\left[\bar{\alpha}^{(t)}\|f(\boldsymbol{\theta}^{(t)})\|^2\right] = \mathcal{O}\bigg(\frac{1}{\eta QT}(f(\boldsymbol{\theta}^{(0)}) - f^*) + G^2$$
$$+ \frac{\eta L c^2}{Q} + L^2\eta^2 Q(\sigma_l^2 + Q\sigma_g^2) + \frac{dL\eta}{n^2 QT}\sum_{t=0}^{T-1} v^{(t)}\bigg), \quad (27)$$

where $f^* = \min_{\boldsymbol{\theta} \in \mathbb{R}^d} f(\boldsymbol{\theta})$, $\bar{\alpha}^{(t)} = (1/n)\sum_{i=1}^n \mathbb{E}[\alpha_i^{(t)}]$ with the expectation taking over mini-batch sampling and $\alpha_i^{(t)} = \min\{1, c/\|\boldsymbol{\Delta}_i^{(t)}\|\}$.

Building upon this foundational lemma, we derive the convergence bound for general AirFL-DP settings.

*Proposition 4.1 (Convergence for AirFL-DP):* Under Assumptions 1, 3–5, AirFL-DP with $\eta = \mathcal{O}(1/(QL))$ after $T$ rounds satisfies

$$\frac{1}{T}\sum_{t=0}^{T-1} \mathbb{E}\left[\bar{\alpha}^{(t)}\|f(\boldsymbol{\theta}^{(t)})\|^2\right]$$
$$= \mathcal{O}\bigg(\frac{1}{\eta QT}(f(\boldsymbol{\theta}^{(0)}) - f^*) + G^2 + L^2\eta^2 Q(\sigma_l^2 + Q\sigma_g^2)$$
$$+ \frac{\eta L c^2}{Q} + \frac{L\eta}{r^2 n^2 QT}\sum_{t=0}^{T-1} \left(d\|\boldsymbol{w}^{(t)}\|^2\sigma^2 + rn\Lambda^{(t)}\right)\bigg), \quad (28)$$

where
$$\Lambda^{(t)} =$$
$$\sum_{j=1}^{d} \mathbb{E}\Bigg[\Bigg|\sum_{i \in \mathcal{I}^{(t)}} \left|(1 - (\boldsymbol{w}^{(t)})^H \boldsymbol{h}_i^{(t)} s_i^{(t)}) \min(\Delta_{i,j}^{(t)}, \frac{c \cdot \Delta_{i,j}^{(t)}}{\|\boldsymbol{\Delta}_i^{(t)}\|})\right|\Bigg|^2\Bigg], \quad (29)$$

and $f^*$, $\bar{\alpha}^{(t)}$, $\alpha_i^{(t)}$ are as defined in Lemma 4.1.

*Proof:* The proof is based on Lemma 4.1 and is detailed in Appendix E. ∎

### B. Problem Formulation

Leveraging Corollary 3.1 and Proposition 4.1, we formulate a learning performance maximization problem. The objective is to minimize the derived convergence upper bound in (28) by optimizing the transceiver design variables $\left\{\{\boldsymbol{w}^{(t)}\}_{t=0}^{T-1}, \{s_i^{(t)}\}_{t=0}^{T-1}\right\}$, subject to transmit power constraint (c.f. (7)) and a privacy level predefined by a constant $\epsilon$. Specifically, the problem is formulated as

$$(\text{P0}): \min_{\{\boldsymbol{w}^{(t)}\}_{t=0}^{T-1}, \{s_i^{(t)}\}_{t=0}^{T-1}} \sum_{t=0}^{T-1} \left(\Lambda^{(t)} + \|\boldsymbol{w}^{(t)}\|^2 d\sigma^2\right)$$
$$\text{s.t.} \quad c^2(s_i^{(t)})^2 \leq dP, \ \forall i \in \mathcal{I}^{(t)}, \ \forall t, \quad (30)$$
$$\epsilon \geq \sqrt{\frac{(2c_\delta + 8)\log(1/\delta)rc^2}{\sigma^2} \min\left\{\sum_{t=0}^{T-1} \phi_t, \Phi\right\}}.$$

Given any $\{\boldsymbol{w}^{(t)}\}_{t=0}^{T-1}$, the optimal $s_i^{(t)}$ is given by [3], [37]

$$s_i^{(t)} = \frac{1}{(\boldsymbol{w}^{(t)})^H \boldsymbol{h}_i^{(t)}}, \tag{31}$$

which yields $\Lambda^{(t)} = 0$, and $\phi_t = 1/\|\boldsymbol{w}^{(t)}\|^2$. By substituting $s_i^{(t)}$, $i \in \mathcal{I}^{(t)}$, $\forall t$, we recast problem (P0) into the following equivalent problem.

$$(\text{P0}'): \min_{\{\boldsymbol{w}^{(t)}\}_{t=0}^{T-1}} \sum_{t=0}^{T-1} \|\boldsymbol{w}^{(t)}\|^2$$

$$\text{s.t.} \quad \left|(\boldsymbol{w}^{(t)})^H \boldsymbol{h}_i^{(t)}\right| \geq \frac{c}{dP}, \ \forall i \in \mathcal{I}^{(t)}, \ \forall t,$$

$$\epsilon \geq \sqrt{\frac{(2c_\delta + 8)\log(1/\delta)rc^2}{\sigma^2} \min\left\{\sum_{t=0}^{T-1} \frac{1}{\|\boldsymbol{w}^{(t)}\|^2}, \Phi\right\}}. \tag{32}$$

For illustration, we solve a simplified version of (P0') assuming an unbounded parameter domain (i.e., $D \to \infty$), yielding $\min\{\sum_t 1/\|\boldsymbol{w}^{(t)}\|^2, \Phi\} = \sum_t 1/\|\boldsymbol{w}^{(t)}\|^2$. Note that the general case with finite $D$ is also tractable and discussed in Sec. VI. In this case, (P0') is simplified as:

$$(\widetilde{\text{P0}}): \min_{\{\boldsymbol{w}^{(t)}\}_{t=0}^{T-1}} \sum_{t=0}^{T-1} \|\boldsymbol{w}^{(t)}\|^2 \tag{33a}$$

$$\text{s.t.} \quad |(\boldsymbol{w}^{(t)})^H \boldsymbol{h}_i^{(t)}| \geq c/\sqrt{dP}, \ \forall i \in \mathcal{I}^{(t)}, \ \forall t, \tag{33b}$$

$$\sum_{t=0}^{T-1} \frac{1}{\|\boldsymbol{w}^{(t)}\|^2} \leq A, \tag{33c}$$

where

$$A = \frac{\epsilon^2 \sigma^2}{(2c_\delta + 8)\log(1/\delta)rc^2}. \tag{34}$$

*Theorem 4.1:* Solving problem $(\widetilde{\text{P0}})$ is equivalent to solving $(\widetilde{\text{P0}}')$, which is defined as

$$(\widetilde{\text{P0}}'): \min_{\{\boldsymbol{w}^{(t)}\}_{t=0}^{T-1}} \sum_{t=0}^{T-1} \|\boldsymbol{w}^{(t)}\|^2 \tag{35a}$$

$$\text{s.t.} \quad \|\boldsymbol{w}^{(t)}\| \geq \|\boldsymbol{w}_{\text{ZF}}^{(t)}\| > 0, \ \forall t, \tag{35b}$$

$$\sum_{t=0}^{T-1} \frac{1}{\|\boldsymbol{w}^{(t)}\|^2} \leq A, \tag{35c}$$

with

$$\boldsymbol{w}_{\text{ZF}}^{(t)} = \frac{c}{\sqrt{dP}} \boldsymbol{H}^{(t)} \left((\boldsymbol{H}^{(t)})^H \boldsymbol{H}^{(t)}\right)^{-1} \boldsymbol{u}, \tag{36}$$

where $\boldsymbol{H}^{(t)} = [\boldsymbol{h}_1^{(t)}, \cdots, \boldsymbol{h}_{rn}^{(t)}] \in \mathbb{C}^{m \times rn}$, and $\boldsymbol{u} \in \mathbb{C}^{rn}$ denotes the vector whose components all have unit modulus.

*Proof:* To prove this result, we first establish the optimal solution in the absence of the DP constraint (33c). Without the DP constraint (33c), $(\widetilde{\text{P0}})$ can be decoupled into $T$ quadratically constrained quadratic programs (QCQPs), each for one $t$ as follows:

$$(\text{P1}): \min_{\boldsymbol{w}^{(t)}} \|\boldsymbol{w}^{(t)}\|^2 \tag{37a}$$

$$\text{s.t.} \quad |(\boldsymbol{w}^{(t)})^H \boldsymbol{h}_i^{(t)}| \geq c/\sqrt{dP}, \ \forall i \in \mathcal{I}^{(t)}, \ \forall t. \tag{37b}$$

Generally, each sub-problem can be individually addressed by using the semidefinite relaxation (SDR) method [38], which involves relaxing the non-convex rank-1 constraint to form a solvable semidefinite program (SDP) [39] and then recovering a rank-1 solution. Here, we provide a more direct analytical solution in the following lemma.

*Lemma 4.2:* Assuming that $m \geq rn$ and i.i.d. channels coefficients, the optimal solution of (P1) is the zero-forcing (ZF) combiner [40]:

$$\boldsymbol{w}_{\text{ZF}}^{(t)} = \frac{c}{\sqrt{dP}} \boldsymbol{H}^{(t)} \left((\boldsymbol{H}^{(t)})^H \boldsymbol{H}^{(t)}\right)^{-1} \boldsymbol{u}, \tag{38}$$

where $\boldsymbol{H}^{(t)} = [\boldsymbol{h}_1^{(t)}, \cdots, \boldsymbol{h}_{rn}^{(t)}] \in \mathbb{C}^{m \times rn}$, and $\boldsymbol{u} \in \mathbb{C}^{rn}$ denotes the vector whose components all have unit modulus.

The detailed proof of Lemma 4.2 is provided in Appendix F. We now prove that solving $(\widetilde{\text{P0}})$ and $(\widetilde{\text{P0}}')$ are equivalent.

Denote the optimal solutions of $(\widetilde{\text{P0}})$ and $(\widetilde{\text{P0}}')$ as $\{\hat{\boldsymbol{w}}^{(t)}\}$ and $\{\widetilde{\boldsymbol{w}}^{(t)}\}$, respectively. First, we note that the solution $\{\hat{\boldsymbol{w}}^{(t)}\}$ also satisfies the constraint of (35b) and (35c) by Lemma 4.2, making it a feasible solution for $(\widetilde{\text{P0}}')$. We then have

$$\sum_{t=0}^{T-1} \|\widetilde{\boldsymbol{w}}^{(t)}\|^2 \leq \sum_{t=0}^{T-1} \|\hat{\boldsymbol{w}}^{(t)}\|^2. \tag{39}$$

In addition, for the solution $\{\frac{\|\widetilde{\boldsymbol{w}}^{(t)}\|}{\|\boldsymbol{w}_{\text{ZF}}^{(t)}\|} \boldsymbol{w}_{\text{ZF}}^{(t)}\}_{t=0}^{T-1}$, we have

$$\left|\left(\frac{\|\widetilde{\boldsymbol{w}}^{(t)}\|}{\|\boldsymbol{w}_{\text{ZF}}^{(t)}\|} \boldsymbol{w}_{\text{ZF}}^{(t)}\right)^H \boldsymbol{h}_i^{(t)}\right| \geq |(\boldsymbol{w}_{\text{ZF}}^{(t)})^H \boldsymbol{h}_i^{(t)}| \geq c/\sqrt{dP}, \quad \forall i \in \mathcal{I}^{(t)}, \tag{40}$$

and

$$\sum_{t=0}^{T-1} \frac{1}{\|\frac{\|\widetilde{\boldsymbol{w}}^{(t)}\|}{\|\boldsymbol{w}_{\text{ZF}}^{(t)}\|} \boldsymbol{w}_{\text{ZF}}^{(t)}\|^2} \leq \sum_{t=0}^{T-1} \frac{1}{\|\widetilde{\boldsymbol{w}}^{(t)}\|^2} \leq A, \tag{41}$$

which means that $\{\frac{\|\widetilde{\boldsymbol{w}}^{(t)}\|}{\|\boldsymbol{w}_{\text{ZF}}^{(t)}\|} \boldsymbol{w}_{\text{ZF}}^{(t)}\}_{t=0}^{T-1}$ is a feasible solution for $(\widetilde{\text{P0}})$. Further, we have

$$\sum_{t=0}^{T-1} \|\hat{\boldsymbol{w}}^{(t)}\|^2 \leq \sum_{t=0}^{T-1} \|\frac{\|\widetilde{\boldsymbol{w}}^{(t)}\|}{\|\boldsymbol{w}_{\text{ZF}}^{(t)}\|} \boldsymbol{w}_{\text{ZF}}^{(t)}\|^2 \leq \sum_{t=0}^{T-1} \|\widetilde{\boldsymbol{w}}^{(t)}\|^2. \tag{42}$$

Combing (39) and (42), we conclude that $\sum_{t=0}^{T-1} \|\hat{\boldsymbol{w}}^{(t)}\|^2 = \sum_{t=0}^{T-1} \|\widetilde{\boldsymbol{w}}^{(t)}\|^2$, which completes the proof. ∎

### C. Optimal Solution

In this subsection, we proceed to derive the optimal solution of problem $(\widetilde{\text{P0}}')$. Let $\pi_t = \|\boldsymbol{w}_{\text{ZF}}^{(t)}\|$ and define $q_t = \|\boldsymbol{w}^{(t)}\|$. Then, we can rewrite $(\widetilde{\text{P0}}')$ as

$$(\widetilde{\text{P0}}''): \min_{\{q_t\}_{t=0}^{T-1}} \sum_{t=1}^{T-1} q_t^2 \tag{43a}$$

$$\text{s.t.} \quad q_t \geq \pi_t > 0, \ \forall t, \tag{43b}$$

$$\sum_{t=0}^{T-1} \frac{1}{q_t^2} \leq A, \tag{43c}$$

which is convex and satisfies Slater's condition. Hence, the Karush-Kuhn-Tucker (KKT) conditions are necessary and sufficient for optimality. The Lagrangian of $(\widetilde{\text{P0}}'')$ can be written as

$$\mathcal{L}\left(\{q_t\}_{t=0}^{T-1}, \mu, \{\nu_t\}_{t=0}^{T-1}\right) = \sum_{t=1}^{T-1} q_t^2 + \mu \left(\sum_{t=0}^{T-1} \frac{1}{q_t^2} - A\right) + \sum_{t=0}^{T-1} \nu_t \left(\pi_t - q_t\right), \quad (44)$$

where $\mu \geq 0$ and $\nu_t \geq 0$ are the Lagrangian multipliers associated with constraints (43b) and (43c), respectively. Further, the KKT conditions are given as

$$\sum_{t=0}^{T-1} \frac{1}{(q_t^*)^2} \leq A, \quad (45a)$$

$$q_t^* \geq \pi_t > 0, \; \forall t, \quad (45b)$$

$$\mu^* \geq 0, \quad (45c)$$

$$\nu_t^* \geq 0, \; \forall t, \quad (45d)$$

$$\left.\frac{\partial \mathcal{L}\left(\{q_t\}, \mu^*, \{\nu_t^*\}\right)}{\partial q_t}\right|_{q_t = q_t^*} = 2q_t^* - \frac{2\mu^*}{(q_t^*)^3} - \nu_t^* = 0, \; \forall t, \quad (45e)$$

$$\mu^* \left(\sum_{t=0}^{T-1} \frac{1}{(q_t^*)^2} - A\right) = 0, \quad (45f)$$

$$\nu_t^*(\pi_t^* - q_t^*) = 0, \; \forall t. \quad (45g)$$

*Proposition 4.2:* The optimal solution of problem $(\widetilde{\text{P0}}'')$ can be expressed as follows:

$$q_t^* = \begin{cases} \pi_t, & \text{if } \sum_{i=0}^{T-1} \frac{1}{\pi_i^2} \leq A \\ \max\left\{\pi_t, (\mu^*)^{\frac{1}{4}}\right\}, & \text{otherwise} \end{cases}, \; \forall t, \quad (46)$$

where $\mu^*$ denotes the optimal value of the dual variable.

*Proof:* First, we solve the KKT conditions by performing a case analysis on the Lagrange multiplier $\mu^*$.

**Case (i)** If $\sum_{t=0}^{T-1} 1/\pi_t^2 \leq A$, the solution $\{\pi_t\}$ satisfies privacy and is optimal, i.e.,

$$q_t^* = \pi_t, \; \forall t \quad (47)$$

**Case (ii)** Otherwise, we have to increase the values of $\{q_t\}$ to satisfy the DP constraint.

**Sub-case (ii-1):** $\mu^* = 0$. From condition (45e), we have $\nu_t^* = 2q_t^*$. Substituting this into conditions (45b) and (45g), we have $q_t^* = \pi_t$ for all $t$. This recovers case (i).

**Sub-case (ii-2):** $\mu^* > 0$. This case requires the privacy constraint to be active, i.e., $\sum_{t=0}^{T-1} 1/(q_t^*)^2 = A$. We then analyze the conditions for each $t$ based on the complementary slackness condition (45g) associated with $\nu_t$:

- If $q_t^* > \pi_t$, then its corresponding slackness condition (45g) requires $\nu_t^* = 0$, which yields $q_t^* = (\mu^*)^{\frac{1}{4}}$ according to condition (45e).
- Else if $q_t^* = \pi_t$, then we only require $\nu_t^* \geq 0$, which leads to $\mu^* \leq \pi_t^4$ according to condition (45e). This is consistent with the initial premise, as $q_t^* = \pi_t \geq (\mu^*)^{\frac{1}{4}}$.

Combining these two sub-cases gives a closed-form expression for the optimal $q_t^*$ in terms of $\mu^*$:

$$q_t^* = \max\left\{\pi_t, (\mu^*)^{\frac{1}{4}}\right\}, \; \forall t. \quad (48)$$

The remaining problem is to determine the value of the Lagrange multiplier $\mu^*$. From previous analysis, the solution for $\{q_t^*\}_{t=0}^{T-1}$ must satisfy the following condition:

$$\sum_{t=0}^{T-1} \frac{1}{q_t^2} = A. \quad (49)$$

By substituting the derived expression $q_t = \max\{\pi_t, (\mu^*)^{\frac{1}{4}}\}$, we obtain an equation solely in terms of $\mu^*$, as follows

$$h(\mu^*) \triangleq \sum_{t=0}^{T-1} \frac{1}{\left[\max\{\pi_t, (\mu^*)^{\frac{1}{4}}\}\right]^2} = A. \quad (50)$$

We note that $h(\mu^*)$ is a monotonically decreasing function of $\mu^*$. This property allows us to efficiently search the unique root of $h(\mu^*) = A$ by using the bisection method. To do so, we establish a search interval $[\mu_l, \mu_h]$ such that $h(\mu_l) > A$ and $h(\mu_h) < A$.

- Determine $\mu_l$. We simply set $\mu_l = 0$, since $h(0) = \sum_{t=0}^{T-1} \frac{1}{\pi_t^2} > A$ as mentioned before.
- Determine $\mu_h$. We need a sufficiently large $\mu_h$ to ensure $h(\mu_h) < A$. A value of $\mu_h$ that satisfies both $\mu_h \geq \max_t \pi_t^4$ and $\mu_h > (\frac{T}{A})^2$ will suffice. The first condition ensures $\mu_h^{\frac{1}{4}}$ dominates $\pi_t$, simplifying $h(\mu_h)$ to $\frac{T}{\sqrt{\mu_h}}$, while the second ensures $\frac{T}{\sqrt{\mu_h}} < A$. Therefore, a valid $\mu_h$ can be constructed as $\mu_h = 1.1 \max\{\max_t \pi_t^4, (\frac{T}{A})^2\}$.

This completes the proof. ∎

*Corollary 4.1:* The optimal solution of problem $(\widetilde{\text{P0}})$ can be expressed as follows:

$$\boldsymbol{w}^{(t)*} = \begin{cases} \boldsymbol{w}_{\text{ZF}}^{(t)}, & \text{if } \sum_{i=0}^{T-1} \frac{1}{\pi_i^2} \leq A \\ \frac{q_t^*}{\pi_t} \boldsymbol{w}_{\text{ZF}}^{(t)}, & \text{otherwise} \end{cases}, \; \forall t, \quad (51)$$

where $q_t^*$ denotes the optimal solution of $(\widetilde{\text{P0}}'')$.

*Remark 4.1 (Overall approach for solving optimization problem):* The objective of $(\widetilde{\text{P0}})$, $\sum_{t=0}^{T-1} \|\boldsymbol{w}^{(t)}\|^2$, depends solely on the norms of the combining vectors $\{\boldsymbol{w}^{(t)}\}$, not their directions. This enables reformulating $(\widetilde{\text{P0}})$ as an optimization over the norms $\{q_t\}_{t=0}^{T-1}$, where $q_t = \|\boldsymbol{w}^{(t)}\|$, subject to feasibility constraints derived from the power constraint (33b) and the DP constraint (33c). For each $t$, the minimum feasible $q_t$ is the norm $\pi_t \triangleq \|\boldsymbol{w}_{\text{ZF}}^{(t)}\|$ of the ZF solution to (P1) in (36). Thus, any valid $q_t$ must satisfy $q_t \geq \pi_t$. This reduces $(\widetilde{\text{P0}})$ to $(\widetilde{\text{P0}}'')$, which optimizes $q_t$ subject to $q_t \geq \pi_t$ for all $t$ and the DP constraint. Once the optimal norm $q_t^*$ is found, we construct an optimal solution via scaling, i.e., $\boldsymbol{w}^{(t)*} = (q_t^*/\pi_t)\boldsymbol{w}_{\text{ZF}}^{(t)}$, which satisfies the power constraint (33b) due to homogeneity.

*Remark 4.2 (When DP can be achieved as a perk):* The optimal solution of $(\widetilde{\text{P0}})$ can be summarized as follows:

- The ZF solution $\{\boldsymbol{w}_{\text{ZF}}^{(t)}\}_{t=0}^{T-1}$ is optimal if $\sum_{t=0}^{T} 1/\pi_t^2 \leq A$, i.e., the maximum signal to noise ratio (SNR) satisfies the following inequality

$$\text{SNR} \triangleq \frac{P}{\sigma^2} \leq \frac{\epsilon^2}{(2c_\delta + 8)\log(1/\delta)rdh_{\text{eff}}}, \quad (52)$$

where the effective channel is defined as

$$h_{\text{eff}} \triangleq \sum_{t=0}^{T-1} \frac{1}{\|\boldsymbol{H}^{(t)}\left((\boldsymbol{H}^{(t)})^H \boldsymbol{H}^{(t)}\right)^{-1} \boldsymbol{u}\|^2}. \quad (53)$$

In this case, *DP can be achieved as a perk by leveraging channel impairments, without compromising training performance.*
- Otherwise, the scaled ZF solution $\{(q_t^*/\pi_t)\boldsymbol{w}_{\text{ZF}}^{(t)}\}_{t=0}^{T-1}$ is optimal, with $q_t^*$ obtained by (46).

*Remark 4.3 (The property of privacy-for-free is also holds in multi-user SIMO system):* Our results enable a *privacy-for-free* mechanism, countering the claims in [26] that zero-artificial-noise is not always possible in DP-guaranteed multi-antenna AirFL systems. Specifically, our design eliminates artificial noise in such systems. In low-SNR regimes meeting the condition (52), the DP-aware solution matches ZF, achieving privacy guarantees without performance loss.

Furthermore, it is worth noting that the core findings of our work exhibit a high degree of generality. While related works, such as [26], may adopt different system models—e.g., variations in the specific form of the DP constraint (arising from different privacy quantification methods or threat models) and the power constraint (due to different communication models)—the fundamental procedure for deriving the optimal solution and the structure of the solution itself remain unaffected. Consequently, *the key insights of our paper, including the privacy-for-free mechanism and the characterization of the utility-privacy trade-off, are preserved* even when adapting our framework to these variations.

## V. EXPERIMENTS

In this section, we present numerical experiments to corroborate our theoretical analyses for privacy and convergence of the AirFL-DP training protocol, as well as verifying in practical settings the range of prescribed privacy level and received SNR where DP can be gained as a perk.

### A. Experimental Setting

We evaluate our protocol on the Fashion-MNIST dataset [41] for a 10-class image classification task. The dataset, comprising $60,000$ training and $10,000$ test images (each $28 \times 28$ pixels), is partitioned in an i.i.d. manner across the clients. Each client holds a disjoint subset of the training data, equally balanced across all classes. The clients train a common CNN model which consists of three convolutional layers and a softmax output layer, totaling $d = 582,026$ parameters. All reported results are averaged over 10 Monte Carlo trials to ensure statistical significance.

Unless otherwise specified, we use the following parameters for our experiments. We simulate the FL system with $n = 50$ WDs and a BS equipped with $m = 100$ antennas. In each of the $T = 50$ communication rounds, all clients participate (client sampling rate $r = 1$) and perform $Q = 5$ local SGD steps with a mini-batch size of 10. The learning rate is set to $\eta = 0.005$, the clipping threshold $c = \sqrt{0.012d}$, and the transmit power is $P = 2$ mW for all devices. For the DP concern, we set the parameters to $\tilde{\epsilon} \triangleq \epsilon/\sqrt{d} = 0.1$ and $\delta = 10^{-5}$.

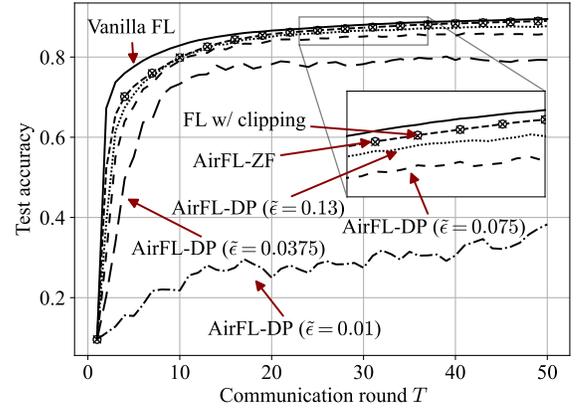

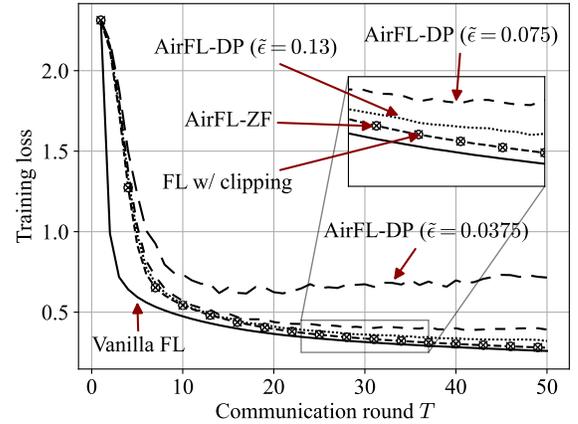

Fig. 3. The evolution of the (a) testing accuracy and (b) training loss versus the global communication rounds $T$ with different privacy budgets $\tilde{\epsilon}$.

We adopt a block-flat Rayleigh fading channel, i.e., $\boldsymbol{h}_i^{(t)} \sim \mathcal{CN}(\boldsymbol{0}, \Lambda \boldsymbol{I})$. The path loss coefficient $\Lambda$ is determined by a simplified path loss model, i.e., $\Lambda = (\frac{c_l}{4\pi f_c r_i})^2$, where $c_l$ denotes the speed of light, $f_c = 2.4$ GHz is the central frequency, and the device-to-BS distance $r_i \sim 1000 \times \sqrt{\mathcal{U}(0,1)}$. Furthermore, the power spectrum density is set to $N_0 = -173$ dBm/Hz over a bandwidth of 20 MHz, resulting in the channel noise scale $\sigma^2 \approx -100$ dBm.

We consider the following benchmarks:

1) Vanilla FL: This scheme refers to the standard FL algorithm implemented over an ideal, noiseless communication channel. It serves as a performance upper bound, disregarding any wireless communication impairments.
2) FL w/ clipping: This scheme augments the vanilla FL with gradient clipping, which is also implemented over an ideal, noiseless channel.
3) AirFL-ZF: This scheme employs ZF beamforming (36) to maximize performance over the fading channel without considering the DP constraint.

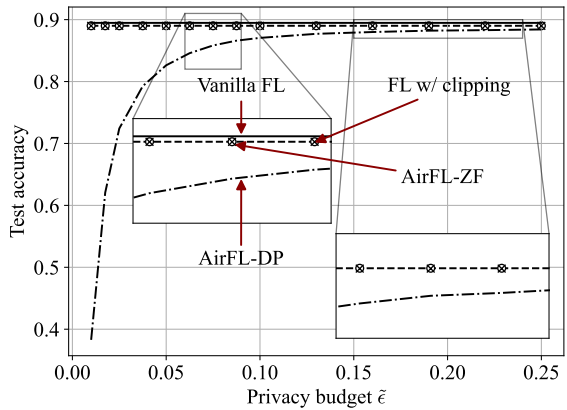

(a)

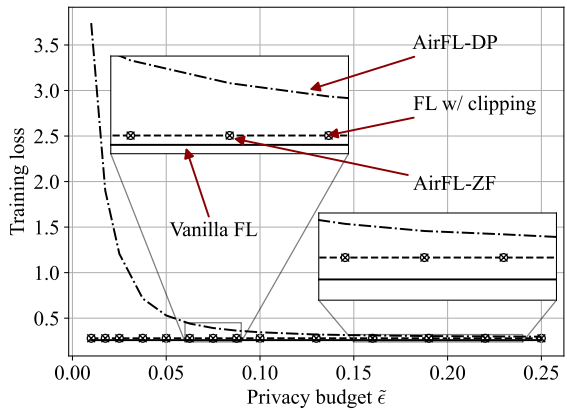

(b)

Fig. 4. The evolution of the (a) testing accuracy and (b) training loss versus the privacy budget $\tilde{\epsilon}$. Note that benchmarks relate to the case of $\tilde{\epsilon} \to \infty$ and are drawn in the same figure for illustration purposes only.

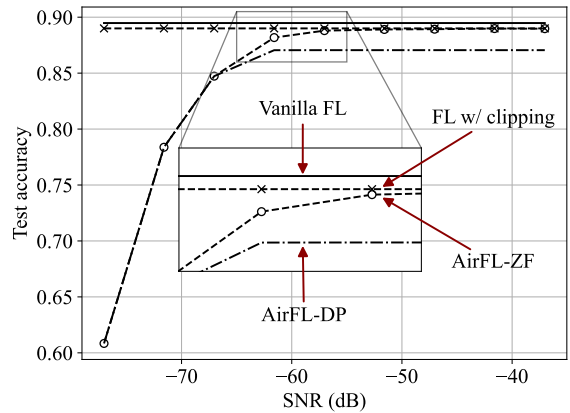

(a)

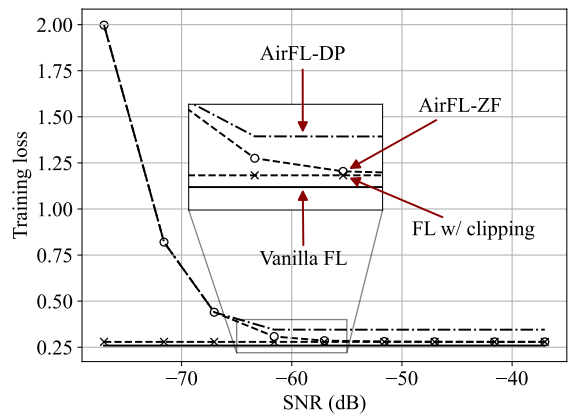

(b)

Fig. 5. The evolution of the (a) testing accuracy and (b) training loss versus the worst-case SNR, $P\Lambda_{\min}/\sigma^2$, where $\Lambda_{\min} = (c_l/(4\pi f_c r_{\max}))^2$ and $r_{\max} = 1000$. The SNR is varied by adjusting the transmit power $P$.

## B. Experimental Analysis

In Fig. 3, we report the test accuracy and training loss versus the communication round $T$ with different privacy budgets $\tilde{\epsilon}$, respectively. The results reveal the inherent privacy-convergence trade-off, where stricter privacy guarantees result in lower test accuracy and higher training loss. Notably, as the privacy budget is relaxed (e.g., to $\tilde{\epsilon} = 0.13$, the performance gap between the AirFL-DP and AirFL-ZF significantly decreases, indicating that AirFL-DP achieves privacy protection for free without performance degradation for a relatively strict privacy level. Furthermore, we observe that the performance of AirFL-ZF is highly comparable to that of FL w/ clipping operating on a perfect channel. It confirms that the ZF solution can effectively mitigate the wireless channel impairments, demonstrating its optimality.

Fig. 4 shows the test accuracy and training loss versus the privacy budget $\tilde{\epsilon}$, where a small $\tilde{\epsilon}$ corresponds to a more stringent privacy requirement. The performance of AirFL-DP sharply approaches the benchmarks as $\tilde{\epsilon}$ increases, demonstrating that AirFL-DP provides free DP guarantees without sacrificing the performance in a wide range of privacy budgets, thanks to the tighter privacy bound and optimal transceiver design.

Finally, Fig. 5 validates our theoretical analysis on the impact of the SNR. This result reveals a distinct threshold effect, as shown in (52). In the low-SNR regime, the performance of AirFL-DP is identical to that of AirFL-ZF, while it performs slightly poorer than AirFL-ZF as SNR increases, which coincides with the observation in Sec. IV.

## VI. CONCLUSIONS AND DISCUSSIONS

In this paper, we studied user-level DP in AirFL systems, considering a multi-antenna base station and multi-access fading channels. We addressed the challenge of harnessing inherent channel noise for privacy—a problem where prior art is often limited by simplified channel models, restrictive assumptions on loss functions, and non-convergent privacy bounds. Our main theoretical contributions were the derivation of a tight, convergent bound on the DP loss under a general bounded-domain assumption, alongside a convergence analysis for smooth, non-convex loss functions. To this end, we optimized the receive beamforming and power allocation to characterize the optimal convergence-privacy trade-off. This analysis demonstrated, contrary to recent claims, that injecting artificial noise is not necessary. Instead, we revealed the explicit conditions under which DP can be achieved as a perk by leveraging channel impairments, without compromising training performance. Finally, the validity of our theoretical findings was confirmed by extensive numerical results.

Lastly, we briefly discussed the solution to the general problem (P0′) with a finite $D$, where the privacy constraint includes the term $\min\{\sum_t \phi_t, \Phi\}$. As mentioned, this problem remains tractable and can be solved using the same principles as the simplified case. A standard approach to handle the min operator is to introduce an auxiliary constraint, which can be incorporated into the Lagrangian function via an additional dual variable. This, in turn, necessitates a subsequent case analysis based on whether the new multiplier is active or inactive. While the resulting derivation becomes more algebraically involved, the core solution approach—rooted in the KKT conditions—remains unchanged. For brevity, we omitted the full derivation, as it follows the same logical steps and offers no new conceptual insights.

## APPENDIX A
## IMPORTANT LEMMAS

*Lemma A.1 (Equivalent Definitions of $\infty$-Wasserstein Distance [30]):* The following are equivalent for any distributions $\mathbb{P}_{\boldsymbol{\mu}}$ and $\mathbb{P}_{\boldsymbol{\nu}}$:
1) $W_\infty(\mathbb{P}_{\boldsymbol{\mu}}, \mathbb{P}_{\boldsymbol{\nu}}) \leq s$.
2) There exist jointly distributed random variables $(\boldsymbol{u}, \boldsymbol{v})$ such that $\boldsymbol{u} \sim \mathbb{P}_{\boldsymbol{\mu}}$, $\boldsymbol{v} \sim \mathbb{P}_{\boldsymbol{\nu}}$, and $\Pr[\|\boldsymbol{u} - \boldsymbol{v}\| \leq s] = 1$.
3) There exist jointly distributed random variables $(\boldsymbol{u}, \boldsymbol{w})$ such that $\boldsymbol{u} \sim \mathbb{P}_{\boldsymbol{\mu}}$, $\boldsymbol{u} + \boldsymbol{w} \sim \mathbb{P}_{\boldsymbol{\nu}}$, and $\Pr[\|\boldsymbol{w}\| \leq s] = 1$.

*Lemma A.2 (Non-Expansiveness of Clipping):* For any $\boldsymbol{x}, \boldsymbol{y} \in \mathbb{R}^d$, we have
$$\|\operatorname{clip}_c(\boldsymbol{x}) - \operatorname{clip}_c(\boldsymbol{y})\| \leq \|\boldsymbol{x} - \boldsymbol{y}\|. \tag{54}$$

*Lemma A.3 (Partial Convexity in the Second Argument [42]):* Let $\boldsymbol{\mu}$, $\boldsymbol{\nu}_1$, and $\boldsymbol{\nu}_2$ be three random variables. Then, for any $\alpha \geq 0$, the Rényi divergence is convex in its second argument, that is, the following inequality holds
$$\begin{aligned}&\mathcal{D}_\alpha(\mathbb{P}_{\boldsymbol{\mu}} \| (1-\lambda)\mathbb{P}_{\boldsymbol{\nu}_1} + \lambda \mathbb{P}_{\boldsymbol{\nu}_2}) \\ &\leq (1-\lambda)\mathcal{D}_\alpha(\mathbb{P}_{\boldsymbol{\mu}} \| \mathbb{P}_{\boldsymbol{\nu}_1}) + \lambda \mathcal{D}_\alpha(\mathbb{P}_{\boldsymbol{\mu}} \| \mathbb{P}_{\boldsymbol{\nu}_2}),\end{aligned} \tag{55}$$
for any $0 \leq \lambda \leq 1$.

*Lemma A.4 (Post-Processing Inequality [42]):* Let $\boldsymbol{\mu}$ and $\boldsymbol{\nu}$ be two random variables. Then, for any (possibly random) function $\psi$ and $\alpha \geq 0$, we have
$$\mathcal{D}_\alpha(\mathbb{P}_{\psi(\boldsymbol{\mu})} \| \mathbb{P}_{\psi(\boldsymbol{\nu})}) \leq \mathcal{D}_\alpha(\mathbb{P}_{\boldsymbol{\mu}} \| \mathbb{P}_{\boldsymbol{\nu}}). \tag{56}$$

*Lemma A.5 (Strong Composition for Rényi Divergence [33]):* Let $\mathcal{M} = [\mathcal{M}_1(\mathcal{D}), \mathcal{M}_2(\mathcal{D})]$ be a sequence of two randomized mechanisms. Then, for any $\alpha > 1$, we have
$$\begin{aligned}\mathcal{D}_\alpha(\mathbb{P}_{\mathcal{M}(\mathcal{D})} \| \mathbb{P}_{\mathcal{M}(\mathcal{D}')}) &\leq \mathcal{D}_\alpha(\mathbb{P}_{\mathcal{M}_1(\mathcal{D})} \| \mathbb{P}_{\mathcal{M}_1(\mathcal{D}')}) \\ &+ \sup_{\boldsymbol{v}} \mathcal{D}_\alpha(\mathbb{P}_{\mathcal{M}_2(\mathcal{D})|\mathcal{M}_1(\mathcal{D})=\boldsymbol{v}} \| \mathbb{P}_{\mathcal{M}_2(\mathcal{D}')|\mathcal{M}_1(\mathcal{D}')=\boldsymbol{v}}),\end{aligned} \tag{57}$$
where $\mathcal{D}$ and $\mathcal{D}'$ are two adjacent datasets.

*Lemma A.6:* For any vector $\boldsymbol{x}$ and $\boldsymbol{y}$, it holds that
$$\boldsymbol{x}^T \boldsymbol{y} = -\frac{1}{2}\|\boldsymbol{x}\|^2 - \frac{1}{2}\|\boldsymbol{y}\|^2 + \frac{1}{2}\|\boldsymbol{x} + \boldsymbol{y}\|^2. \tag{58}$$

*Lemma A.7:* For any $a, b \in \mathbb{R}$, it holds that
$$(a + b)^2 \leq 2a^2 + 2b^2. \tag{59}$$

*Lemma A.8:* For any learning rate $\eta$ satisfying $\eta \leq \frac{1}{8QL}$, under Assumptions 1 and 4-5, we can bound the drift for any $q \in \{0, \ldots, Q-1\}$ as
$$\begin{aligned}&\frac{1}{n}\sum_{i=1}^n \mathbb{E}\left\|\boldsymbol{\theta}_i^{(t,q)} - \boldsymbol{\theta}^{(t)}\right\|^2 \\ &\leq 5Q\eta^2\left(\sigma_l^2 + 6Q\sigma_g^2\right) + 30Q^2\eta^2 \|\nabla f(\boldsymbol{\theta}^{(t)})\|^2.\end{aligned} \tag{60}$$

## APPENDIX B
## PROOF OF LEMMA 3.3

*Proof:* We first consider the special case of a single local update, i.e., $Q = 1$. Based on Lemma A.1, for the shifted Rényi divergence $\mathcal{D}_\alpha^{(z)}(\mathbb{P}_{\boldsymbol{\mu}} \| \mathbb{P}_{\boldsymbol{\nu}})$, there exist jointly distributed random variables $(\boldsymbol{\mu}, \boldsymbol{\mu}')$ such that $\Pr[\|\boldsymbol{\mu} - \boldsymbol{\mu}'\| \leq z] = 1$ and $\mathcal{D}_\alpha^{(z)}(\mathbb{P}_{\boldsymbol{\mu}} \| \mathbb{P}_{\boldsymbol{\nu}}) = \mathcal{D}_\alpha(\mathbb{P}_{\boldsymbol{\mu}'} \| \mathbb{P}_{\boldsymbol{\nu}})$. For this one-step mapping, we have
$$\begin{aligned}&\|\psi^{(t)}(\boldsymbol{\mu}) - \psi^{(t)}(\boldsymbol{\mu}')\| \\ &\leq \|\boldsymbol{\mu} - \boldsymbol{\mu}'\| + \frac{\eta L}{rn}\sum_{i \in \mathcal{I}^{(t)}}(\boldsymbol{w}^{(t)})^H \boldsymbol{h}_i^{(t)} s_i^{(t)} \|\boldsymbol{\mu} - \boldsymbol{\mu}'\| \\ &\leq (1 + \frac{\eta L}{rn}\sum_{i \in \mathcal{I}^{(t)}}(\boldsymbol{w}^{(t)})^H \boldsymbol{h}_i^{(t)} s_i^{(t)}) z \\ &= (1 + \kappa^{(t)}) z,\end{aligned} \tag{61}$$
with
$$\kappa^{(t)} = \frac{\eta L}{rn}\sum_{i \in \mathcal{I}^{(t)}}(\boldsymbol{w}^{(t)})^H \boldsymbol{h}_i^{(t)} s_i^{(t)}, \tag{62}$$
where the first step is by the triangle inequality, Assumption 1, and Lemma A.2, and the second step is by the definition of $\boldsymbol{\mu}'$. When round $t$ involves $Q$ successive local SGD updates, the overall mapping $\psi^{(t)}$ can be viewed as the composition of $Q$ such one-step mappings. Then, the $Q$-step bound immediately follows as
$$\|\psi^{(t)}(\boldsymbol{\mu}) - \psi^{(t)}(\boldsymbol{\mu}')\| \leq (1 + \kappa^{(t)})^Q z. \tag{63}$$

Thus, we have
$$\begin{aligned}&\mathcal{D}_\alpha^{((1+\kappa^{(t)})^Q z)}(\mathbb{P}_{\psi^{(t)}(\boldsymbol{\mu})} \| \mathbb{P}_{\psi^{(t)\prime}(\boldsymbol{\nu})}) \\ &\leq \mathcal{D}_\alpha(\mathbb{P}_{\psi^{(t)}(\boldsymbol{\mu}')} \| \mathbb{P}_{\psi^{(t)\prime}(\boldsymbol{\nu})}) \\ &= \mathcal{D}_\alpha(\mathbb{P}_{\psi^{(t)}(\boldsymbol{\mu}')} \| \mathbb{P}_{(1-r)\psi^{(t)}(\boldsymbol{\nu}) + r\psi^{(t)\prime\prime}(\boldsymbol{\nu})}) \\ &\leq (1-r)\mathcal{D}_\alpha(\mathbb{P}_{\psi^{(t)}(\boldsymbol{\mu}')} \| \mathbb{P}_{\psi^{(t)}(\boldsymbol{\nu})}) + r\mathcal{D}_\alpha(\mathbb{P}_{\psi^{(t)}(\boldsymbol{\mu}')} \| \mathbb{P}_{\psi^{(t)\prime\prime}(\boldsymbol{\nu})}) \\ &\leq (1-r)\mathcal{D}_\alpha(\mathbb{P}_{\boldsymbol{\mu}'} \| \mathbb{P}_{\boldsymbol{\nu}}) + r\mathcal{D}_\alpha(\mathbb{P}_{\psi^{(t)}(\boldsymbol{\mu}')} \| \mathbb{P}_{\psi^{(t)\prime\prime}(\boldsymbol{\nu})}),\end{aligned} \tag{64}$$
where the first step is by Definition 3.2, the second step is due to device sampling with $\psi''(\cdot) \triangleq \psi'(\cdot | i^* \in \mathcal{I}^{(t)})$, the third step is by Lemma A.3, and the last step is by Lemma A.4.

As for the second term $\mathcal{D}_\alpha(\mathbb{P}_{\psi^{(t)}(\boldsymbol{\mu}')}||\mathbb{P}_{\psi^{(t)''}(\boldsymbol{\nu})})$, we have

$$\begin{aligned}&\mathcal{D}_\alpha(\mathbb{P}_{\psi^{(t)}(\boldsymbol{\mu}')}||\mathbb{P}_{\psi^{(t)''}(\boldsymbol{\nu})})\\&\leq \mathcal{D}_\alpha(\mathbb{P}_{\psi^{(t)}(\boldsymbol{\mu}'),\boldsymbol{\mu}'}||\mathbb{P}_{\psi^{''(t)}(\boldsymbol{\nu}),\boldsymbol{\nu}})\\&\leq \sup_{\boldsymbol{v}} \mathcal{D}_\alpha(\mathbb{P}_{\psi^{(t)}(\boldsymbol{\mu}')|\boldsymbol{\mu}'=\boldsymbol{v}}||\mathbb{P}_{\psi^{(t)''}(\boldsymbol{\nu})|\boldsymbol{\nu}=\boldsymbol{v}}) + \mathcal{D}_\alpha(\mathbb{P}_{\boldsymbol{\mu}'}||\mathbb{P}_{\boldsymbol{\nu}})\\&\leq \frac{2\alpha c^2 \max_{i\in\mathcal{I}^{(t)}}|(\boldsymbol{w}^{(t)})^H \boldsymbol{h}_i^{(t)} s_i^{(t)}|^2}{\beta^{(t)}\|\boldsymbol{w}^{(t)}\|^2 \sigma^2} + \mathcal{D}_\alpha(\mathbb{P}_{\boldsymbol{\mu}'}||\mathbb{P}_{\boldsymbol{\nu}}),\end{aligned}$$
(65)

where the first step is by Lemma A.4, the second step is by the Lemma A.5, and the last step is by the well-known result $\mathcal{D}_\alpha(\mathcal{N}(\boldsymbol{0},\sigma^2 \boldsymbol{I}_d)||\mathcal{N}(\boldsymbol{u},\sigma^2 \boldsymbol{I}_d)) = \alpha\|\boldsymbol{u}\|_2^2/2\sigma^2$. Hence,

$$\begin{aligned}&\mathcal{D}_\alpha^{((1+\kappa^{(t)})^Q z)}(\mathbb{P}_{\psi^{(t)}(\boldsymbol{\mu})}||\mathbb{P}_{\psi^{(t)'}(\boldsymbol{\nu})})\\&\leq \mathcal{D}_\alpha^{(z)}(\mathbb{P}_{\boldsymbol{\mu}}||\mathbb{P}_{\boldsymbol{\nu}}) + \frac{2\alpha rc^2 \max_{i\in\mathcal{I}^{(t)}}|(\boldsymbol{w}^{(t)})^H \boldsymbol{h}_i^{(t)} s_i^{(t)}|^2}{\beta^{(t)}\|\boldsymbol{w}^{(t)}\|^2 \sigma^2},\end{aligned}$$
(66)

which completes the proof. ∎

## APPENDIX C
## PROOF OF PROPOSITION 3.1

*Proof:* Let $\boldsymbol{\theta}^{(T)}$ and $\boldsymbol{\theta}'^{(T)}$ denote the output of AirFL-DP based on user-adjacent datasets $\mathcal{D}$ and $\mathcal{D}' = \mathcal{D} \cup \mathcal{D}_{i^*}$, respectively. Let $\Xi_\tau = \max_{\mathcal{D},\mathcal{D}'} \|\boldsymbol{\theta}^{(\tau)} - \boldsymbol{\theta}'^{(\tau)}\|$ denote the maximum parameter perturbation at round $\tau$ resulting from two adjacent datasets. Define

$$\kappa_{\max} = \max_t \kappa^{(t)} = \max_t \frac{\eta L}{rn} \sum_{i\in\mathcal{I}^{(t)}} (\boldsymbol{w}^{(t)})^H \boldsymbol{h}_i^{(t)} s_i^{(t)}. \quad (67)$$

Then, we consider a real sequence $\{a_k\}_{k=\tau}^{T-1}$ and any $\tau \in \{0,1,\cdots,T-1\}$ such that $z_t = ((1+\kappa_{\max})^Q)^{t-\tau}\Xi_\tau - \sum_{k=\tau}^{t-1}((1+\kappa_{\max})^Q)^{t-k-1}a_k$ is non-negative for all $t \geq \tau$ and $z_T = 0$. By this way, we have $z_\tau = \Xi_\tau$ and $z_{t+1} = (1+\kappa_{\max})^Q z_t - a_t$. Furthermore, the proof is conducted by induction, utilizing Lemma 3.2 and Lemma 3.3. Specifically, we have

$$\begin{aligned}&\mathcal{D}_\alpha^{(z_{t+1})}(\mathbb{P}_{\boldsymbol{\theta}^{(t+1)}}||\mathbb{P}_{\boldsymbol{\theta}'^{(t+1)}})\\&= \mathcal{D}_\alpha^{(z_{t+1})}(\mathbb{P}_{\psi^{(t)}(\boldsymbol{\theta}^{(t)})+\boldsymbol{\varsigma}^{(t)}}||\mathbb{P}_{\psi^{(t)'}(\boldsymbol{\theta}'^{(t)})+\boldsymbol{\varsigma}^{(t)}})\\&\leq \mathcal{D}_\alpha^{(z_{t+1}+a_t)}(\mathbb{P}_{\psi^{(t)}(\boldsymbol{\theta}^{(t)})}||\mathbb{P}_{\psi^{(t)'}(\boldsymbol{\theta}'^{(t)})}) + \frac{\alpha a_t^2 r^2 n^2}{2(1-\beta^{(t)})W^{(t)}}\\&= \mathcal{D}_\alpha^{((1+\kappa_{\max})^Q z_t)}(\mathbb{P}_{\psi^{(t)}(\boldsymbol{\theta}^{(t)})}||\mathbb{P}_{\psi^{(t)'}(\boldsymbol{\theta}'^{(t)})}) + \frac{\alpha a_t^2 r^2 n^2}{2(1-\beta^{(t)})W^{(t)}}\\&\leq \mathcal{D}_\alpha^{((1+\kappa^{(t)})^Q z_t)}(\mathbb{P}_{\psi^{(t)}(\boldsymbol{\theta}^{(t)})}||\mathbb{P}_{\psi^{(t)'}(\boldsymbol{\theta}'^{(t)})}) + \frac{\alpha a_t^2 r^2 n^2}{2(1-\beta^{(t)})W^{(t)}}\\&\leq \mathcal{D}_\alpha^{(z_t)}(\mathbb{P}_{\boldsymbol{\theta}^{(t)}}||\mathbb{P}_{\boldsymbol{\theta}'^{(t)}}) + \frac{2\alpha rc^2 \phi_t}{\beta^{(t)}\sigma^2} + \frac{\alpha a_t^2 r^2 n^2}{2(1-\beta^{(t)})W^{(t)}},\end{aligned}$$
(68)

with

$$W_t = \eta^2 \sigma^2 \|\boldsymbol{w}^{(t)}\|^2$$
$$\phi_t = \frac{\max_{i\in\mathcal{I}^{(t)}}|(\boldsymbol{w}^{(t)})^H \boldsymbol{h}_i^{(t)} s_i^{(t)}|^2}{\|\boldsymbol{w}^{(t)}\|^2}, \quad (69)$$

where the second step is by Lemma 3.2, the fourth step is by the Definition 3.2, and the last step is by Lemma 3.3. By repeating the induction from $T$ to $\tau$, we can obtain

$$\begin{aligned}\mathcal{D}_\alpha(\mathbb{P}_{\boldsymbol{\theta}^{(T)}}||\mathbb{P}_{\boldsymbol{\theta}'^{(T)}}) &\leq \mathcal{D}_\alpha^{(z_\tau)}(\mathbb{P}_{\boldsymbol{\theta}^{(\tau)}}||\mathbb{P}_{\boldsymbol{\theta}'^{(\tau)}})\\&+ \sum_{t=\tau}^{T-1}\frac{2\alpha rc^2 \phi_t}{\beta^{(t)}\sigma^2} + \sum_{t=\tau}^{T-1}\frac{\alpha a_t^2 r^2 n^2}{2\eta^2 \sigma^2 (1-\beta^{(t)})\|\boldsymbol{w}^{(t)}\|^2}.\end{aligned}$$
(70)

Note that $\mathcal{D}_\alpha^{(z_\tau)}(\mathbb{P}_{\boldsymbol{\theta}^{(\tau)}}||\mathbb{P}_{\boldsymbol{\theta}'^{(\tau)}}) = 0$. Let $a_\tau = (1+\kappa_{\max})^Q \Xi_\tau$ and $a_t = 0$ for all $t > \tau$, one may obtain

$$\mathcal{D}_\alpha(\mathbb{P}_{\boldsymbol{\theta}^{(T)}}||\mathbb{P}_{\boldsymbol{\theta}'^{(T)}}) \leq \sum_{t=\tau}^{T-1}\frac{2\alpha rc^2 \phi_t}{\beta^{(t)}\sigma^2} + \frac{\alpha(1+\kappa_{\max})^{2Q}\Xi_\tau^2 r^2 n^2}{2\eta^2 \sigma^2 (1-\beta^{(\tau)})\|\boldsymbol{w}^{(\tau)}\|^2} \quad (71)$$

for any $\tau \in \{0,\ldots,T-1\}$ and $\beta^{(t)} \in [0,1]$. Concerning $\Xi_\tau$, we have

$$\begin{cases}\Xi_\tau = 0, & \text{if } \tau = 0,\\ \Xi_\tau \leq D, & \text{if } \tau > 0.\end{cases} \quad (72)$$

We then derive the privacy bound by performing a case analysis on the auxiliary round $\tau$.

**Case 1:** $\tau = 0$. In this case, we have

$$\mathcal{D}_\alpha(\mathbb{P}_{\boldsymbol{\theta}^{(T)}}||\mathbb{P}_{\boldsymbol{\theta}'^{(T)}}) \leq \sum_{t=0}^{T-1}\frac{2\alpha rc^2 \phi_t}{\sigma^2}. \quad (73)$$

**Case 2:** $\tau > 0$. In this case, we have

$$\mathcal{D}_\alpha(\mathbb{P}_{\boldsymbol{\theta}^{(T)}}||\mathbb{P}_{\boldsymbol{\theta}'^{(T)}}) \leq \min_{\tau\in\{0,1,\ldots,T-1\},\{\beta^{(t)}\}}\left\{\sum_{t=\tau}^{T-1}\frac{2\alpha rc^2 \phi_t}{\beta^{(t)}\sigma^2} + \frac{\alpha(1+\kappa_{\max})^{2Q}D^2 r^2 n^2}{2\eta^2 \sigma^2 (1-\beta^{(\tau)})\|\boldsymbol{w}^{(\tau)}\|^2}\right\}. \quad (74)$$

Hence, we can obtain

$$\begin{aligned}&\mathcal{D}_\alpha(\mathbb{P}_{\boldsymbol{\theta}^T}||\mathbb{P}_{\boldsymbol{\theta}'^{(T)}})\\&\leq \min_{\beta^{(T-1)}}\frac{2\alpha rc^2 \phi_{T-1}}{\beta^{(T-1)}\sigma^2} + \frac{\alpha(1+\kappa_{\max})^{2Q}D^2 r^2 n^2}{2\eta^2 \sigma^2 (1-\beta^{(T-1)})\|\boldsymbol{w}^{(T-1)}\|^2}\\&\leq \left(\sqrt{\frac{2\alpha rc^2 \phi_{T-1}}{\sigma^2}} + \sqrt{\frac{\alpha(1+\kappa_{\max})^{2Q}D^2 r^2 n^2}{2\eta^2 \sigma^2 \|\boldsymbol{w}^{(T-1)}\|^2}}\right)^2\\&= \frac{2\alpha rc^2}{\sigma^2}\Phi.\end{aligned}$$
(75)

with

$$\Phi = \left(\sqrt{\phi_{T-1}} + \frac{(1+\kappa_{\max})^Q \sqrt{r}Dn}{2\eta c\|\boldsymbol{w}^{(T-1)}\|}\right)^2. \quad (76)$$

In summary, we can obtain

$$\begin{aligned}\mathcal{D}_\alpha(\mathbb{P}_{\boldsymbol{\theta}^{(T)}}||\mathbb{P}_{\boldsymbol{\theta}'^{(T)}}) &\leq \min\left\{\sum_{t=0}^{T-1}\frac{2\alpha rc^2 \phi_t}{\sigma^2}, \frac{2\alpha rc^2}{\sigma^2}\Phi\right\}\\&\leq \frac{2\alpha rc^2}{\sigma^2}\min\left\{\sum_{t=0}^{T-1}\phi_t, \Phi\right\},\end{aligned}$$
(77)

which completes the proof. ∎

# APPENDIX D
## PROOF OF COROLLARY 3.1

*Proof:* Recall that

$$\mathcal{D}_\alpha(\mathbb{P}_{\boldsymbol{\theta}^T} || \mathbb{P}_{\boldsymbol{\theta}'^{(T)}}) \leq \frac{2\alpha r c^2}{\sigma^2} \min\left\{\sum_{t=0}^{T-1} \phi_t, \Phi\right\}. \quad (78)$$

Let $\epsilon > 0$ and $0 < \delta < 1$ be two constants such that $\epsilon' \leq \frac{c_\delta}{4}\log(1/\delta)$. To transform $(\alpha, \epsilon')$-RDP into the standard $(\epsilon, \delta)$-DP characterization, we use Lemma 3.1 by setting $\alpha = 1 + \frac{2}{\epsilon}\log(1/\delta)$ and $\epsilon' = \epsilon/2$, obtaining

$$\epsilon = \sqrt{\frac{(2c_\delta + 8)\log(1/\delta)rc^2}{\sigma^2} \min\left\{\sum_{t=0}^{T-1} \phi_t, \Phi\right\}}, \quad (79)$$

which completes the proof. ∎

# APPENDIX E
## PROOF OF PROPOSITION 4.1

*Proof:* Beginning with Assumption 1 and taking expectation conditioned on $\boldsymbol{\theta}^{(t)}$, we have

$$\mathbb{E}\left[f(\boldsymbol{\theta}^{(t+1)})\right] \leq f(\boldsymbol{\theta}^{(t)}) + \frac{L}{2}\eta^2 c^2 + \frac{L}{2}\frac{\eta^2}{r^2 n^2}\mathbb{E}\left[\|\boldsymbol{n}_{\text{est}}^{(t)}\|^2\right]$$
$$+ \mathbb{E}\left[\nabla f(\boldsymbol{\theta}^{(t)})^T\left(-\frac{\eta}{n}\sum_{i=1}^n \text{clip}_c(\boldsymbol{\Delta}_i^{(t)})\right)\right]. \quad (80)$$

(1) For term $\mathbb{E}\left[\nabla f(\boldsymbol{\theta}^{(t)})^T\left(-\frac{\eta}{n}\sum_{i=1}^n \text{clip}_c(\boldsymbol{\Delta}_i^{(t)})\right)\right]$:

Define $\alpha_i^{(t)} = \min\{1, \frac{c}{\|\boldsymbol{\Delta}_i^{(t)}\|}\}$ and $\bar{\alpha}^{(t)} = \frac{1}{n}\sum_{i=1}^n \mathbb{E}[\alpha_i^{(t)}]$. Then, we can obtain

$$\mathbb{E}\left[\nabla f(\boldsymbol{\theta}^{(t)})^T\left(-\frac{\eta}{n}\sum_{i=1}^n \text{clip}_c(\boldsymbol{\Delta}_i^{(t)})\right)\right]$$
$$= \nabla f(\boldsymbol{\theta}^{(t)})^T \mathbb{E}\left[-\frac{\eta}{n}\sum_{i=1}^n (\alpha_i^{(t)}\boldsymbol{\Delta}_i^{(t)} - \bar{\alpha}^{(t)}\boldsymbol{\Delta}_i^{(t)} + \bar{\alpha}^{(t)}\boldsymbol{\Delta}_i^{(t)})\right]$$
$$\leq \eta Q G^2 + \nabla f(\boldsymbol{\theta}^{(t)})^T \mathbb{E}\left[-\frac{\eta}{n}\sum_{i=1}^n \bar{\alpha}^{(t)}\boldsymbol{\Delta}_i^{(t)}\right], \quad (81)$$

where the last step is by the Cauchy-Schwarz inequality.

(1.1) For term $\nabla f(\boldsymbol{\theta}^{(t)})^T \mathbb{E}\left[-\frac{\eta}{n}\sum_{i=1}^n \bar{\alpha}^{(t)}\boldsymbol{\Delta}_i^{(t)}\right]$:

$$\nabla f(\boldsymbol{\theta}^{(t)})^T \mathbb{E}\left[-\frac{\eta}{n}\sum_{i=1}^n \bar{\alpha}^{(t)}\boldsymbol{\Delta}_i^{(t)}\right]$$
$$\leq -\frac{1}{2}\eta\bar{\alpha}^{(t)}Q\|\nabla f(\boldsymbol{\theta}^{(t)})\|^2 - \frac{\eta^2}{2}\frac{\bar{\alpha}^{(t)}}{\eta Q}\|\frac{1}{n}\sum_{i=1}^n \mathbb{E}[\boldsymbol{\Delta}_i^{(t)}]\|^2 +$$
$$\frac{1}{2}\mathbb{E}\left[\|\sqrt{\eta\bar{\alpha}^{(t)}Q}\nabla f(\boldsymbol{\theta}^{(t)}) - \frac{\eta}{\sqrt{\eta\bar{\alpha}^{(t)}Q}}\frac{1}{n}\sum_{i=1}^n \bar{\alpha}^{(t)}\mathbb{E}[\boldsymbol{\Delta}_i^{(t)}]\|^2\right]$$
$$\leq -\frac{1}{2}\eta\bar{\alpha}^{(t)}Q\|\nabla f(\boldsymbol{\theta}^{(t)})\|^2$$
$$+ \frac{\eta\bar{\alpha}^{(t)}Q}{2}\mathbb{E}\left[\|\nabla f(\boldsymbol{\theta}^{(t)}) - \frac{1}{Q}\frac{1}{n}\sum_{i=1}^n \mathbb{E}[\boldsymbol{\Delta}_i^{(t)}]\|^2\right], \quad (82)$$

where the first step is by Lemma A.6.

(1.2) For term $\mathbb{E}\left[\|\nabla f(\boldsymbol{\theta}^{(t)}) - \frac{1}{Q}\frac{1}{n}\sum_{i=1}^n \mathbb{E}[\boldsymbol{\Delta}_i^{(t)}]\|^2\right]$:

$$\mathbb{E}\left[\|\nabla f(\boldsymbol{\theta}^{(t)}) - \frac{1}{Q}\frac{1}{n}\sum_{i=1}^n \mathbb{E}[\boldsymbol{\Delta}_i^{(t)}]\|^2\right]$$
$$\leq \frac{1}{n}\sum_{i=1}^n \mathbb{E}\left[\|\nabla f_i(\boldsymbol{\theta}^{(t)}) - \frac{1}{Q}\mathbb{E}[\boldsymbol{\Delta}_i^{(t)}]\|^2\right]$$
$$\leq \frac{2}{n}\sum_{i=1}^n \mathbb{E}\left[\|\frac{1}{Q}\sum_{q=0}^{Q-1}\left(\nabla f_i(\boldsymbol{\theta}^{(t)}) - \nabla f_i(\boldsymbol{\theta}_i^{(t,q)})\right)\|^2\right] \quad (83)$$
$$+ \frac{2}{n}\sum_{i=1}^n \mathbb{E}\left[\|\frac{1}{Q}\sum_{q=0}^{Q-1}\nabla f_i(\boldsymbol{\theta}_i^{(t,q)}) - \frac{1}{Q}\mathbb{E}[\boldsymbol{\Delta}_i^{(t)}]\|^2\right]$$
$$\leq \frac{2}{n}\sum_{i=1}^n \frac{1}{Q}\sum_{q=0}^{Q-1} L^2 \mathbb{E}\left[\|\boldsymbol{\theta}^{(t)} - \boldsymbol{\theta}_i^{(t,q)}\|^2\right],$$

where the first step is by the Jensen's inequality, the second step is by Lemma A.7, and the last step is by Assumption 1.

Using Lemma A.8, we can derive that

$$\mathbb{E}\left[\|\nabla f(\boldsymbol{\theta}^{(t)}) - \frac{1}{Q}\frac{1}{n}\sum_{i=1}^n \mathbb{E}[\boldsymbol{\Delta}_i^{(t)}]\|^2\right]$$
$$\leq \frac{2}{n}\sum_{i=1}^n L^2\left(5Q\eta^2(\sigma_l^2 + 6Q\sigma_g^2) + 30Q^2\eta^2\|\nabla f(\boldsymbol{\theta}^{(t)})\|^2\right)$$
$$= 10L^2 Q\eta^2(\sigma_l^2 + 6Q\sigma_g^2) + 60L^2 Q^2 \eta^2 \|\nabla f(\boldsymbol{\theta}^{(t)})\|^2. \quad (84)$$

(2) For term $\mathbb{E}\left[\|\boldsymbol{n}_{\text{est}}^{(t)}\|^2\right]$:

Recall that the $j$-th entry of the estimation error $\boldsymbol{n}_{\text{est}}^{(t)}$ can be expressed as $n_{\text{est},j}^{(t)} = \left(\hat{\Delta}_j^{(t)} - \Delta_j^{(t)}\right)$. Define

$$\Lambda^{(t)} = \sum_{j=1}^d \mathbb{E}\left[|\sum_{i \in \mathcal{I}^{(t)}}(1 - (\boldsymbol{w}^{(t)})^H \boldsymbol{h}_i^{(t)} s_i^{(t)})\bar{\Delta}_{i,j}^{(t)}|^2\right]$$
$$\leq \sum_{j=1}^d \mathbb{E}\left[\sum_{i \in \mathcal{I}^{(t)}}\left|(1 - (\boldsymbol{w}^{(t)})^H \boldsymbol{h}_i^{(t)} s_i^{(t)})\min(\Delta_{i,j}^{(t)}, \frac{c \cdot \Delta_{i,j}^{(t)}}{\|\boldsymbol{\Delta}_i^{(t)}\|})\right|^2\right], \quad (85)$$

yielding

$$\mathbb{E}\left[\|\boldsymbol{n}_{\text{est}}^{(t)}\|^2\right] = \Lambda^{(t)} + \|\boldsymbol{w}^{(t)}\|^2 d\sigma^2 \quad (86)$$

(3) In summary, we have

$$\mathbb{E}\left[f(\boldsymbol{\theta}^{(t+1)})\right]$$
$$\leq f(\boldsymbol{\theta}^{(t)}) + \eta Q G^2 - \frac{1}{2}\eta\bar{\alpha}^{(t)}Q\|\nabla f(\boldsymbol{\theta}^{(t)})\|^2 + \frac{L}{2}\eta^2 c^2$$
$$+ \frac{\eta\bar{\alpha}^{(t)}Q^2 L^2}{2}\left(10\eta^2(\sigma_l^2 + 6Q\sigma_g^2) + 60Q\eta^2\|\nabla f(\boldsymbol{\theta}^{(t)})\|^2\right)$$
$$+ \frac{L}{2}\frac{\eta^2}{r^2 n^2}\left(\Lambda^{(t)} + \|\boldsymbol{w}^{(t)}\|^2 d\sigma^2\right). \quad (87)$$

Let $\eta \leq \frac{1}{\sqrt{120QL}}$

$$\mathbb{E}\left[f(\boldsymbol{\theta}^{(t+1)})\right] \leq f(\boldsymbol{\theta}^{(t)}) + \eta QG^2 - \frac{1}{4}\eta\bar{\alpha}^{(t)}Q\|\nabla f(\boldsymbol{\theta}^{(t)})\|^2 \\ + 5L^2\eta^3\bar{\alpha}^{(t)}Q^2\left(\sigma_l^2 + 6Q\sigma_g^2\right) + \frac{L}{2}\eta^2 c^2 \\ + \frac{L}{2}\frac{\eta^2}{r^2 n^2}\left(\Lambda^{(t)} + \|\boldsymbol{w}^{(t)}\|^2 d\sigma^2\right). \quad (88)$$

Summing over $t$ from 0 to $T-1$, divide both sides by $\frac{\eta QT}{4}$, take expectation over all, and rearrange, we have

$$\frac{1}{T}\sum_{t=0}^{T-1}\mathbb{E}\left[\bar{\alpha}^{(t)}\|f(\boldsymbol{\theta}^{(t)})\|^2\right] \\ \leq \frac{4}{\eta QT}(f(\boldsymbol{\theta}^{(0)}) - f^*) + \frac{2\eta Lc^2}{Q} + 20L^2\eta^2 Q(\sigma_l^2 + 6Q\sigma_g^2) \\ + \frac{4L\eta}{r^2 n^2 Q}\frac{1}{T}\sum_{t=0}^{T-1}\Lambda^{(t)} + \frac{4dL\eta\sigma^2}{r^2 n^2 Q}\frac{1}{T}\sum_{t=0}^{T-1}\|\boldsymbol{w}^{(t)}\|^2 + 4G^2, \quad (89)$$

which completes the proof. ∎

## APPENDIX F
## PROOF OF LEMMA 4.2

*Proof:* Let $\boldsymbol{H}^{(t)} = [\boldsymbol{h}_1^{(t)}, \cdots, \boldsymbol{h}_{rn}^{(t)}] \in \mathbb{C}^{m \times rn}$. Then, the original optimization problem (P1) can be reformulated as

$$(\text{P1}') : \min_{\boldsymbol{w}^{(t)}} \quad \|\boldsymbol{w}^{(t)}\|^2 \quad (90\text{a})$$
$$\text{s.t.} \quad |(\boldsymbol{H}^{(t)})^H \boldsymbol{w}^{(t)}| \geq \frac{c}{\sqrt{dP}}\mathbf{1}. \quad (90\text{b})$$

Note that both the objective function (90a) and the constraint (90b) are invariant to phase shifts of the optimization variable $\boldsymbol{w}^{(t)}$. Assuming the channel vectors are i.i.d. and $m \geq rn$, we can rotate the phase of $\boldsymbol{w}^{(t)}$ to obtain the following real-form problem, i.e.,

$$(\widetilde{\text{P1}}) : \min_{\boldsymbol{w}^{(t)}} \quad \|\boldsymbol{w}^{(t)}\|^2 \quad (91\text{a})$$
$$\text{s.t.} \quad (\boldsymbol{H}^{(t)})^H \boldsymbol{w}^{(t)} \geq \frac{c}{\sqrt{dP}}\mathbf{1}. \quad (91\text{b})$$

This problem is convex and satisfies Salter's condition. Therefore, the Karush-Kuhn-Tucker (KKT) conditions are necessary and sufficient for optimality. The Lagrangian of $(\widetilde{\text{P1}})$ is given by

$$\mathcal{L}\left(\boldsymbol{w}^{(t)}, \boldsymbol{\lambda}^{(t)}\right) \\ = [\boldsymbol{w}^{(t)}]^H \boldsymbol{w}^{(t)} + [\boldsymbol{\lambda}^{(t)}]^T\left(\frac{c}{\sqrt{dP}}\mathbf{1} - [\boldsymbol{H}^{(t)}]^H \boldsymbol{w}^{(t)}\right), \quad (92)$$

where $\boldsymbol{\lambda}^{(t)}$ is the Lagrangian multiplier associated with constraint (91b). The corresponding KKT conditions are given as

$$[\boldsymbol{H}^{(t)}]^H \boldsymbol{w}^{(t)*} \geq \frac{c}{\sqrt{dP}}\mathbf{1}, \quad (93\text{a})$$
$$\boldsymbol{\lambda}^{(t)*} \geq \mathbf{0}, \quad (93\text{b})$$
$$\left.\frac{\partial \mathcal{L}\left(\boldsymbol{w}^{(t)}, \boldsymbol{\lambda}^{(t)*}\right)}{\partial \boldsymbol{w}^{(t)}}\right|_{\boldsymbol{w}_t = \boldsymbol{w}_t^*} = 2\boldsymbol{w}^{(t)*} - \boldsymbol{H}^{(t)}\boldsymbol{\lambda}^{(t)*} = 0, \quad (93\text{c})$$
$$\boldsymbol{\lambda}_i^{(t)*}\left(\frac{c}{\sqrt{dP}} - [\boldsymbol{h}_i^{(t)}]^H \boldsymbol{w}^{(t)*}\right) = 0, \ \forall i \in \mathcal{I}^{(t)}. \quad (93\text{d})$$

Substituting (93c) into (93a) yields

$$\frac{1}{2}[\boldsymbol{H}^{(t)}]^H \boldsymbol{H}^{(t)} \boldsymbol{\lambda}^{(t)*} \geq \frac{c}{\sqrt{dP}}\mathbf{1}, \quad (94)$$

which implies

$$\boldsymbol{\lambda}^{(t)*} \geq \frac{2c}{\sqrt{dP}}\left[[\boldsymbol{H}^{(t)}]^H \boldsymbol{H}^{(t)}\right]^{-1}\mathbf{1}. \quad (95)$$

Since $\boldsymbol{H}^{(t)}$ is full column rank and $m > rn$, the matrix $[\boldsymbol{H}^{(t)}]^H \boldsymbol{H}^{(t)}$ is positive-definite. Therefore, we have $\boldsymbol{\lambda}^{(t)*} > \mathbf{0}$. It then follows from condition (93d) that

$$\frac{c}{\sqrt{dP}}\mathbf{1} - [\boldsymbol{H}^{(t)}]^H \boldsymbol{w}^{(t)*} = \mathbf{0} \quad (96)$$

which leads to the optimal solutions

$$\boldsymbol{w}^{(t)*} = \frac{c}{\sqrt{dP}}\boldsymbol{H}^{(t)}\left[[\boldsymbol{H}^{(t)}]^H \boldsymbol{H}^{(t)}\right]^{-1}\mathbf{1}, \\ \boldsymbol{\lambda}^{(t)*} = \frac{2c}{\sqrt{dP}}\left[[\boldsymbol{H}^{(t)}]^H \boldsymbol{H}^{(t)}\right]^{-1}\mathbf{1}. \quad (97)$$

Due to the phase-invariant property of the solutions to (P1), the general form of the optimal solution can be expressed as

$$\boldsymbol{w}_{\text{ZF}}^{(t)} = \frac{c}{\sqrt{dP}}\boldsymbol{H}^{(t)}\left((\boldsymbol{H}^{(t)})^H \boldsymbol{H}^{(t)}\right)^{-1}\boldsymbol{u}, \quad (98)$$

where $\boldsymbol{u} \in \mathbb{C}^{rn}$ denotes the vector whose components all have unit modulus.